\documentclass[10pt,twocolumn,letterpaper,pagenumbers]{article}

\usepackage[pagenumbers]{iccv} 

\usepackage{tcolorbox}
\usepackage{float}
\usepackage{tikz} 
\usepackage{xcolor}

\newcommand{\code}[1]{\texttt{#1}}
\newcommand{\ourmethod}{{{TaxaDiffusion}}\xspace}
\newcommand{\ourmethodG}{{{TaxaGuide}}\xspace}

\newcommand\mypara[1]{\vspace{0.5mm}\noindent\textbf{#1}}

\usepackage{tocloft}

\newcommand{\gradientline}{
    \begin{center}
        \begin{tikzpicture}
            \shade[left color=gray!10, right color=gray!10, middle color=gray!50] 
                (0,0) rectangle (6cm, 0.2mm);
        \end{tikzpicture}
    \end{center}
}

\usepackage{amsmath,amsfonts,bm}

\def\eqref#1{equation~\ref{#1}}

\def\1{\bm{1}}

\DeclareMathAlphabet{\mathsfit}{\encodingdefault}{\sfdefault}{m}{sl}
\SetMathAlphabet{\mathsfit}{bold}{\encodingdefault}{\sfdefault}{bx}{n}

\usepackage{multirow}
\usepackage{xcolor}
\usepackage{svg}

\definecolor{citecolor}{RGB}{34,139,34}

\definecolor{iccvblue}{rgb}{0.21,0.49,0.74}
\usepackage[pagebackref,breaklinks,colorlinks,allcolors=iccvblue]{hyperref}

\def\paperID{5374} 
\def\confName{ICCV}
\def\confYear{2025}

\title{
  \textbf{\textit{
    \textcolor{red}{T}
    \textcolor{orange}{a}
    \textcolor{yellow}{x}
    \textcolor{green}{a}
    \textcolor{cyan}{D}
    \textcolor{blue}{i}
    \textcolor{violet}{f}
    \textcolor{magenta}{f}
    \textcolor{purple}{u}
    \textcolor{red!80!black}{s}
    \textcolor{orange!80!black}{i}
    \textcolor{yellow!50!red}{o}
    \textcolor{green!50!black}{n}
    \textcolor{blue!40!black}{:}
  }}
  Progressively Trained Diffusion Model\\
  for Fine-Grained Species Generation
}

\author{
Amin Karimi Monsefi $^{1}$ \quad
Mridul Khurana $^{2}$ \quad
Rajiv Ramnath $^{1}$ \quad
Anuj Karpatne $^{2}$ \\ [2pt]
Wei-Lun Chao $^{1}$ \quad
Cheng Zhang $^{3}$ \\ [6pt]
{$^{1}$ {The Ohio State University} \quad
$^{2}$ {Virginia Tech} \quad
$^{3}$ {Texas A\&M University}
}
}

\begin{document}
\maketitle
\begin{abstract} 
We propose \ourmethod, a taxonomy-informed training framework for diffusion models to generate fine-grained animal images with high morphological and identity accuracy.
Unlike standard approaches that treat each species as an independent category, \ourmethod incorporates domain knowledge that many species exhibit strong visual similarities, with distinctions often residing in subtle variations of shape, pattern, and color. To exploit these relationships, \ourmethod progressively trains conditioned diffusion models across different taxonomic levels --- starting from broad classifications such as \code{Class} and \code{Order}, refining through \code{Family} and \code{Genus}, and ultimately distinguishing at the \code{Species} level. This hierarchical learning strategy first captures coarse-grained morphological traits shared by species with common ancestors, facilitating knowledge transfer, before refining fine-grained differences for species-level distinction. As a result, \ourmethod enables accurate generation even with limited training samples per species.
Extensive experiments on three fine-grained animal datasets demonstrate that \ourmethod outperforms existing approaches, achieving superior fidelity in fine-grained animal image generation. 
Project page:~\href{https://amink8.github.io/TaxaDiffusion/}{https://amink8.github.io/TaxaDiffusion/}
\end{abstract}

\section{Introduction}
\label{sec:intro}

\begin{figure}[t]
    \centering
    \includegraphics[width=1\linewidth]{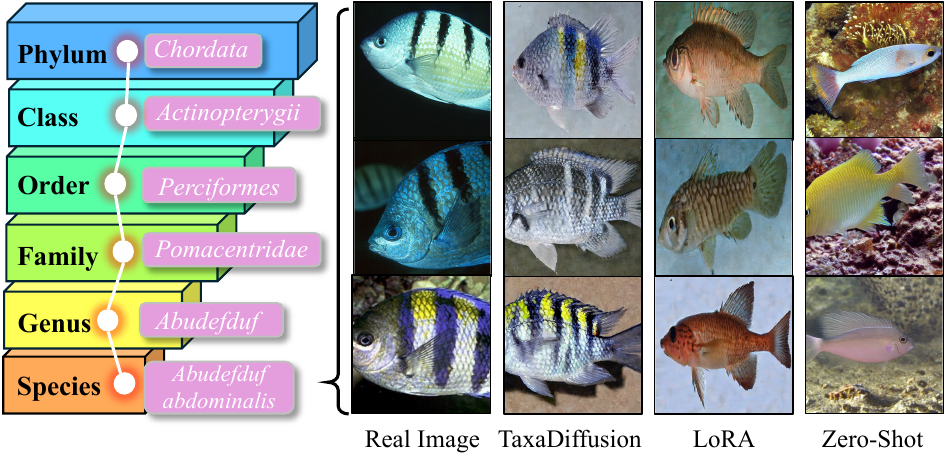}
    \vspace{-7mm}
    \caption{\small Taxonomy encodes a rich hierarchical structure for categorizing life. We propose \ourmethod to leverage such knowledge to enable fine-grained, controllable image generation. Compared to Zero-Shot generation with vanilla Stable Diffusion~\cite{rombach2022high} and LoRA fine-tuning~\cite{hu2021lora}, \ourmethod achieves higher accuracy and captures fine details that align closely with real images.}
    \label{fig:teaser}\vspace{-3mm}
\end{figure}

\begin{figure*}[ht]
    \centering
    \includegraphics[width=1\linewidth]{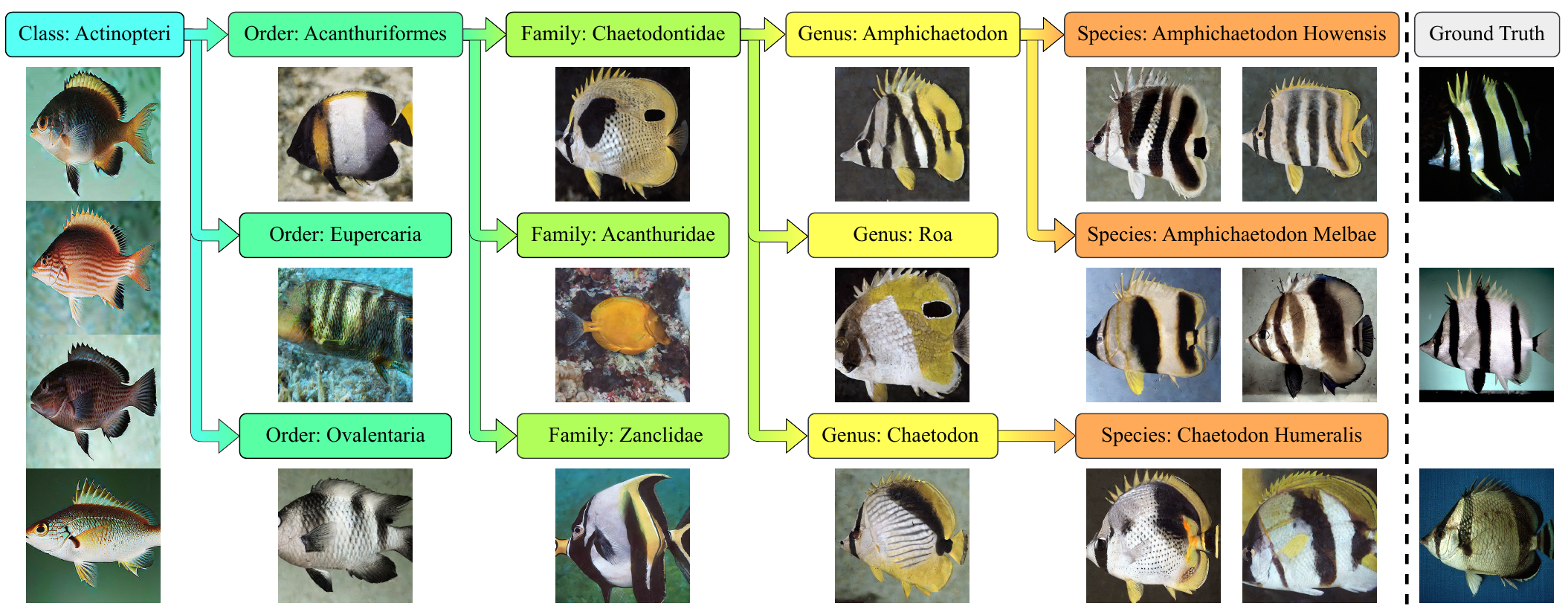}
    \vspace{-5mm}
    \caption{\small \textbf{Generative examples of our approach on the FishNet dataset}~\cite{khan2023fishnet}. As we progress through the taxonomy tree from \code{Class} to \code{Order}, \code{Family}, \code{Genus}, and finally \code{Species}, our model refines its understanding of distinguishing traits, generating realistic images that capture the unique visual characteristics at each level. For rare species with limited training samples, such as ``Amphichaetodon Howensis'' ($4$ samples), ``Amphichaetodon Melbae'' ($1$ sample), and ``Chaetodon Humeralis'' ($5$ samples), our taxonomy-informed, progressive training approach enables effective knowledge transfer from related species, allowing the model to generate morphologically accurate species images even with sparse data. The corresponding ground-truth images from FishNet are shown on the right. 
    } 
    \label{fig:overall}
   \vskip -10pt
\end{figure*}

Diffusion models have revolutionized the field of image generation \cite{podell2023sdxl, ho2020denoising, dhariwal2021diffusion}.
By progressively denoising sampled noise, these models can generate diverse, high-quality images with fine details. Further enhanced by vision-language models \cite{radford2021learning} and various control mechanisms \cite{navard2024knobgen, chan2023generative, zhang2023sine, zhang2023adding}, diffusion models can produce complex compositions of common concepts (e.g., objects and attributes) while faithfully aligning with user inputs.

However, when it comes to generating fine-grained concepts, \emph{specifically animal species}, diffusion models still face significant challenges.
Given a species name, existing models often fail to generate morphologically accurate images or images that precisely reflect the identity of species.

Two key domain characteristics contribute to this issue. First, animals are highly dynamic objects with considerable degrees of freedom. They exhibit various poses and movements, resulting in significant intra-class variation. Second, with millions of animal species on Earth, collecting sufficient samples to capture both intra-class variations and inter-class distinctions is a daunting task. These challenges paint a grim picture for accurate animal generation. We ask:

\begin{center}
\emph{
    How can we learn high-quality generative models for animals without a large amount of data per species? 
}
\end{center}
 
Our key insight is that, despite the vast number of animal species, they are not independent categories. Instead, it took half a billion years for the Animalia Kingdom to evolve from a few species to millions, and species that diverged more recently often share similar visual traits. While CLIP~\cite{radford2021learning, monsefi2024detailclip}, commonly used for prompting in diffusion models, may have learned species relationships from image-text pairs, it primarily relies on correlations. These correlations are only reliable when sufficient data is available --- often not the case for fine-grained animal species.

Therefore, we propose leveraging \textbf{Taxonomy}~\cite{padial2010integrative}, a classification system that has been continuously developed by biologists over decades, if not centuries, to categorize species (Figure \ref{fig:teaser}). A typical taxonomy consists of $7$ hierarchical levels~\cite{dubois2007phylogeny, Hobern2021}: \code{Kingdom}, \code{Phylum}, \code{Class}, \code{Order}, \code{Family}, \code{Genus}, and \code{Species}. Species sharing a common ancestor often exhibit similar characteristics, such as body parts, layouts, and valid poses. The closer two species are on the taxonomy tree, the more similar their appearance tends to be. For species within the same Genus or Family, their distinctions often lie in subtle variations in shape, color, or pattern. To exploit these relationships, we ask:

\begin{center}
\emph{
    How can we use taxonomy in training diffusion models? 
}
\end{center}

One straightforward approach is to replace the common name of a species (e.g, ``cat'') with its full taxonomic hierarchy (e.g., ``Animalia''-``Chordata''-``Mammalia''-``Carnivora''-``Felidae''-``Felis''-``Catus'')~\cite{stevens2024bioclip,khurana2024hierarchical}.
However, introducing all taxonomic levels at once overwhelms the training process, making it hard for the model to effectively learn taxonomic relationships.

To address this issue, we propose a \textbf{progressive} training approach that teaches the model taxonomy step by step, from root to leaf. We hypothesize that this progressive strategy will encourage the model to first learn common body structures, layouts, and poses shared across species, and then refine its focus on subtle differences in shape, color, and pattern --- thus unlocking the full potential of a generative model for synthesizing high-quality animal images. Figure~\ref{fig:overall} presents actual examples generated by our method.

We implement our approach, \textbf{\ourmethod}, as follows, using an animal image dataset labeled with full taxonomic names. Building upon a pre-trained diffusion model (e.g., Stable Diffusion v1-5~\cite{rombach2022high}), we retrain a new condition encoder to process taxonomy information (see Section~\ref{sec:method} for details).
Instead of feeding the full taxonomic name for each image at once, we break the training into multiple progressive stages, each corresponding to a different taxonomy level. At the $i$-th stage, we input the taxonomic name up to the $i$-th level, encouraging the model to learn common visual properties from images labeled with the same taxonomic name up to that level. For example,  ``lion'' and ``hyena'', rivals in many African ecosystems, will be treated as belonging to the same category up to the \code{Order} level, with both labeled as ``Animalia''-``Chordata''-``Mammalia''-``Carnivora.'' As a result, the diffusion model can explicitly learn the common visual characteristics of  ``Carnivora'' from images of many species within this group. 

By progressively refining visual granularity, \ourmethod further enables species with limited data to leverage information from related species, promoting knowledge transfer.  
Specifically, for a rare species, the model only needs to learn how it differs from other species within the same \code{Genus} at the final stage --- usually subtle variations in shape, color, and pattern. Its other visual characteristics can be learned from images of related species in earlier stages.

We empirically validate \ourmethod for fine-grained species generation across three challenging datasets: FishNet~\cite{khan2023fishnet}, BIOSCAN-1M~\cite{gharaee2024step}, and iNaturalist~\cite{van2018inaturalist}, which cover a diverse range of species. Specifically, FishNet~\cite{khan2023fishnet} includes around $100,000$ images spanning over $17,000$ fine-grained fish species, making it a challenging few-shot generation dataset. Extensive experiments and analyses demonstrate that \ourmethod excels in fine-grained, few-shot tasks and opens new avenues for trait discovery in the generative modeling of animal species.

\mypara{Remark.} 
The millions of animal species on Earth did not appear all at once but emerged over half a billion years of evolution, progressively branching out from a single species into many. We aim to teach the diffusion model in a way that mimics this evolutionary process --- \emph{not by learning to generate all species at once, but progressively}. We hypothesize that this approach will enable the model to better learn how to generate animal images. While taxonomy does not perfectly capture the evolutionary process (e.g., genetic similarity), it does capture visual similarity. We hope that our work demonstrates the benefits of incorporating domain knowledge into algorithm design and fosters synergy between the vision and scientific communities.

\section{Related Work}
\label{sec:related::work}
\mypara{Diffusion-based Image Generation.}
Denoising diffusion models have emerged as the leading method for high-quality image generation, surpassing traditional models like GANs \cite{goodfellow2014generative} and VAEs \cite{kingma2013auto} in stability and visual fidelity~\cite{ho2020denoising, dhariwal2021diffusion, ho2021classifierfree, nichol2021glide, saharia2022photorealistic, ramesh2022hierarchical, peebles2023scalable, navard2024knobgen}. While large-scale diffusion models~\cite{ramesh2022hierarchical, saharia2022photorealistic, rombach2022high} excel in generating diverse, high-quality images from captioned datasets~\cite{schuhmann2022laion}, they focus on broad concepts and often lack precision in fine-grained species generation, where subtle distinctions are critical. We address this limitation by enhancing diffusion models to capture intricate details among closely related categories.

\mypara{Fine-grained Generation.}
Current fine-grained generation methods focus on images with subtle distinctions within specific categories, such as fish species, nuanced expressions, or detailed object attributes. This is challenging due to high intra-class similarity and the need to capture distinguishing details~\cite{parihar2025precisecontrol, pan2024finediffusion, matsunaga2022fine, ye2023fine, wang2023fg, zeng2024dilightnet, khurana2024hierarchical,wang2024towards,tan2024empirical}. Existing approaches, including sketch-based control, attribute manipulation, and personalized generation, enhance specificity through attribute conditioning and guidance techniques. Sketch-based control enables detail capture through user sketches \cite{navard2024knobgen, zhao2024uni, li2025controlnet, zhang2023adding}, while attribute manipulation allows precise feature control by conditioning on latent spaces \cite{huang2024diffusion, kim2022diffusionclip, yue2023chatface, han2023highly, matsunaga2022fine,zhang2023iti}. Personalized generation creates context-specific images for individual subjects \cite{zhang2024survey, ruiz2023dreambooth, li2024stylegan, wang2024animatelcm, ruiz2024hyperdreambooth}. Despite their progress, these methods often rely on domain-specific data, lack generalization, or require costly fine-tuning. Our approach addresses these limitations with a progressive, hierarchical training strategy, enabling fine-grained generation across categories without sacrificing detail or necessitating extensive tuning.

\mypara{Progressive Hierarchical Generative Model.}
Hierarchical generative models enhance image quality and scalability \cite{xu2021generative, li2023progressive, pi2023hierarchical}. For instance, \citet{razavi2019generating} introduced a multi-layer autoencoder that captures global structures and fine details by conditioning high-resolution codes on coarse-grained latent samples. Similarly, \cite{child2020very, vahdat2020nvae} used VAEs with progressive latent hierarchies to improve generation efficiency. Hierarchical conditioning on segmentation maps, as in \citet{gafni2022make}, allows semantic map-based image generation. While effective, these methods often encounter reconstruction errors across abstraction levels, reducing fidelity. In diffusion models, hierarchical structures enhance both efficiency and detail capture, as shown in the Hierarchical Integration Diffusion Model (HI-Diff) \cite{chen2024hierarchical}, which integrates prior features in a latent space for image deblurring. However, this regression-based approach may limit novel fine-grained detail generation. This underscores the need for hierarchical frameworks that progressively capture fine details without compromising efficiency or fidelity.

\section{Approach}
\label{sec:method}

\begin{figure*}[t]
    \centering
    \includegraphics[width=1\textwidth]{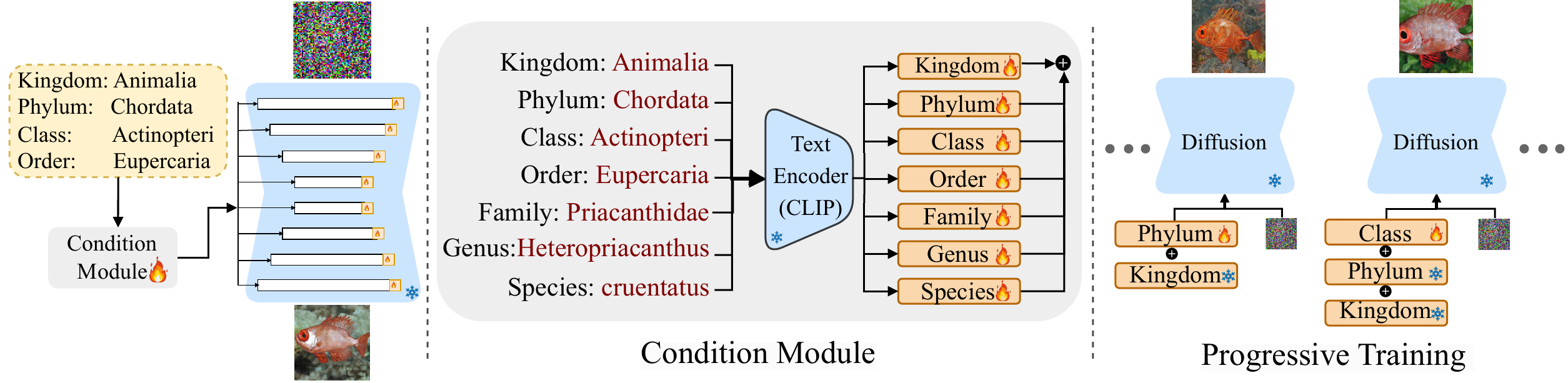}
    \vspace{-6mm}
    \caption{\small \textbf{Overview of \ourmethod for progressive diffusion model training.} \textbf{Left and middle:} We incorporate a taxonomy-informed conditioning module to guide the model training. \textbf{Right:} The hierarchical taxonomy levels (i.e., \code{Kingdom}, \code{Phylum}, \code{Class}, \code{Order}, \code{Family}, \code{Genus}, \code{Species}) are progressively encoded by the text encoder of the CLIP model during training. Each taxonomy level activates specific traits, allowing the model to capture common features at higher levels and fine details at the species level.
    }
    \vspace{-2mm}
    \label{fig:framework}
\end{figure*}

We introduce the background of diffusion models and Taxonomy in Section~\ref{sec:background}, followed by our progressive training approach in Section~\ref{sec:training}. Finally, we describe fine-grained species generation using \ourmethod in Section~\ref{sec:guidance}.  

\subsection{Revisiting Diffusion Models and Taxonomy} \label{sec:background}

\mypara{Diffusion Models.} Conditional diffusion models~\cite{rombach2022high} generate samples conditioned on auxiliary information $\mathbf{c}$ (e.g., class labels, text prompt, or taxonomic levels in this work), enabling attribute-controlled generation. Beginning with data sample $\mathbf{x}_0$, the forward process adds Gaussian noise, producing noisy samples $\mathbf{x}_t$:
\begin{equation} 
\mathbf{x}_t = \sqrt{\bar{\alpha}_t}, \mathbf{x}_0 + \sqrt{1 - \bar{\alpha}_t}, \boldsymbol{\epsilon}, \quad \text{with} \quad \boldsymbol{\epsilon} \sim \mathcal{N}\left( \mathbf{0},, \mathbf{I} \right), 
\end{equation}
where $\alpha_t = 1 - \beta_t$, $\bar{\alpha}_t = \prod_{s=0}^t \alpha_s$, and ${\beta_t}$ a variance schedule at timestep $t$.

The reverse process reconstructs $\mathbf{x}_0$ from noisy $\mathbf{x}_T$ by iteratively denoising with a model conditioned on $\mathbf{c}$, predicting noise added in the forward step:
\begin{equation} 
\mathbf{x}_{t-1} = \frac{1}{\sqrt{\alpha_t}} \left( \mathbf{x}_t - \frac{\beta_t}{\sqrt{1 - \bar{\alpha}_t}}, \boldsymbol{\epsilon}_\theta\left( \mathbf{x}_t, t, \mathbf{c} \right) \right) + \sigma_t \mathbf{z}, \quad 
\end{equation}
\noindent where $\sigma_t$ is typically related to $\beta_t$ and $\mathbf{z} \sim \mathcal{N}\left( \mathbf{0},, \mathbf{I} \right)$. The vanilla model is trained end-to-end for minimizing the difference between the true noise $\boldsymbol{\epsilon}$ and the predicted noise $\boldsymbol{\epsilon}_\theta$:
\begin{equation} \mathcal{L}(\theta) = \mathbb{E}_{\mathbf{x}_0, \boldsymbol{\epsilon}, t} \left[ || \boldsymbol{\epsilon} - \boldsymbol{\epsilon}_\theta\left( \mathbf{x}_t, t, \mathbf{c} \right) || ^2 \right], 
\end{equation}
where $\mathbf{x}_t$ is obtained from $\mathbf{x}_0$ and $\boldsymbol{\epsilon}$ using the forward process, conditioned on $\mathbf{c}$. As discussed in Section~\ref{sec:intro}, using a single $\mathbf{c}$ to handle all taxonomic levels simultaneously can overwhelm the training process, especially for rare species.

\mypara{Classifier-Free Guidance}~\cite{ho2021classifierfree}. During inference, CFG is widely used for improving the generation quality and diversity by refining predictions without a separate classifier \citep{ho2022classifier}. Specifically, CFG combines conditional and unconditional score estimates within one model framework: given a conditional model $\boldsymbol{\epsilon}_\theta(\mathbf{x}_t, t, \mathbf{c})$ and an unconditional version $\boldsymbol{\epsilon}_{\theta}(\mathbf{x}_t, t)$, the adjusted score is:
\begin{equation}\label{eq:cfg}
\tilde{\boldsymbol{\epsilon}}_\theta(\mathbf{x}_t, t, \mathbf{c}) = (1 + w) \times \boldsymbol{\epsilon}_\theta(\mathbf{x}_t, t, \mathbf{c}) - w \times \boldsymbol{\epsilon}_\theta(\mathbf{x}_t, t), 
\end{equation}
where $\boldsymbol{\epsilon}_\theta$ is the noise prediction network and $w$ is the guidance scale. This approach directs samples toward the conditioned distribution without needing a classifier.

\mypara{Taxonomy Levels.} 
We use the index $i$ to denote a specific level within the hierarchical taxonomy, ranging from $0$ to the total number of levels. Notably, $i$ represents the taxonomic rank (\code{Order} or \code{Family}) rather than a numeral value. This differs from the timestep $t$ in the denoising process of diffusion models and from the training epoch.

\subsection{Progressive \ourmethod Training} \label{sec:training}

We propose to progressively train our model to learn from higher taxonomic levels (\code{Kingdom}, \code{Phylum} or \code{Class}) to capture shared traits before refining it to focus on fine-grained, species-specific traits as we advance towards the \code{Species} level in a controlled and efficient manner.
Figure~\ref{fig:framework} illustrates our progressive training process.

\mypara{Efficient Domain Adaptation.}
Pretrained diffusion models, such as Stable Diffusion, are trained on natural images and lack exposure to biology-specific data. Therefore, we first adapt a pretrained diffusion model (Stable Diffusion v1-5) to our dataset using LoRA (Low-Rank Adaptation)~\cite{hu2021lora}. Specifically, LoRA introduces low-rank updates to the self-attention and cross-attention layers, adjusting the query (Q), key (K), value (V), and output projections while keeping the original model weights frozen. The adapted feature representation is formulated as:
\begin{equation} 
\mathbf{Q} = \mathcal{W}^Q \mathbf{z} + \text{AdapterLayer}(\mathbf{z}) = \mathcal{W}^Q \mathbf{z} + \alpha \cdot \mathbf{A} \mathbf{B}^T \mathbf{z},
\end{equation}
\noindent where \( \mathcal{W}^Q \) is the original query projection, $\alpha$ is a scaling factor (defaulted to 1), and $\mathbf{A}$ and $\mathbf{B}$ are learnable low-rank matrices that capture domain-specific knowledge.

However, fine-tuning the model solely with all species labels leads to suboptimal results, as shown in Figure~\ref{fig:teaser} and our experiments. To address this, we introduce a progressive training approach to systematically integrates taxonomy knowledge into the generative model.

\mypara{Taxonomy-Informed Progressive Training.}
Building on the LoRA fine-tuned model\footnote{LoRA is trained only at the first level (\code{Kingdom}) and remains frozen during the subsequent progressive training.}(Figure~\ref{fig:framework}-left), we use a frozen CLIP encoder to transform textual descriptions of each taxonomic level into latent representations. These representations are then refined by a trainable module with two transformer layers (Figure~\ref{fig:framework}-middle). Each module receives output embeddings from the CLIP-text encoder, maintaining a consistent embedding structure of up to 77 tokens (standard CLIP setting of Stable Diffusion 1.5). These embeddings flow consistently through the module, ensuring stable integration across taxonomy levels into the diffusion model.

Our progressive training process proceeds hierarchically, starting from the broadest level (\code{Kingdom}) and gradually refining down to the finest level (\code{Species}). Each conditioning module is trained separately at its respective level\footnote{We train each level for 250K iterations. See appendix for more details.}. Once a level is trained, the corresponding parameters (i.e., transformer layers) are frozen before progressing to the next level (Figure~\ref{fig:framework}-right). 
At each stage, we aggregate embeddings across all trained levels by adding the output embeddings and feeding this combined signal into the diffusion model, as shown in Figure~\ref{fig:framework}-middle. This enables the model to generate images that integrate multi-level taxonomic knowledge --- capturing broad features at higher levels and fine-grained, species-specific details at lower levels. Notably, only the modules corresponding to the current level remain active, ensuring that prior knowledge is retained and stabilized while preventing interference from untrained modules. This sequential freezing mechanism establishes a structured cascade of conditioning signals, progressively enhancing the ability of the model to generate accurate species traits across taxonomic levels.

\subsection{Fine-grained Species Generation}\label{sec:guidance}
As discussed in Section~\ref{sec:background}, CFG generates images by combining conditional and unconditional score estimates. In our case, the conditional representation $\mathbf{c}^{(i)}$ corresponds to the target species at taxonomy $i$. Unlike standard CFG, we investigate a new way that fully leverages higher-level taxonomy information to guide fine-grained species generation.

Specifically, instead of using an unconditional estimate for guidance, we replace it with the conditioning signal from a higher taxonomic level (i.e., \code{Kingdom}) to guide species generation at a given level $i$. That is, the original score estimate in CFG (Equation~\ref{eq:cfg}) can be modified as: 
\begin{equation} \label{eq:taxadiffG}
(1 + w) \times \boldsymbol{\epsilon}_\theta(\mathbf{x}_t, t, {\color{blue}{\mathbf{c}^{(i)}}}) - w \times \boldsymbol{\epsilon}_\theta(\mathbf{x}_t, t, {\color{blue}{\mathbf{c}^{(0)}}}), 
\end{equation}
where $\mathbf{c}^{(i)}$ is the conditioning signal at the target level, and $\mathbf{c}^{(0)}$ is the conditioning from \code{Kingdom}. This reformulation ensures that the guidance is informed by higher taxonomic levels, refining species-specific generation while maintaining shared characteristics inherited from broader levels.

More generally speaking, the final guidance aligns with the \ourmethod training process, facilitating nuanced visual exploration by incorporating hierarchical knowledge from ancestor levels. This idea is reminiscent of the recent work by Karras et al.~\cite{karras2024guiding}, which shows that guiding a diffusion model using a weaker version of itself can improve the quality. Similarly, in our case, high-level conditions serve as a broad domain prior for fine-grained species generation.

\section{Experiments}
\label{sec:experiment}

We begin by introducing the experimental setup in Section~\ref{sec:setup}, then present the main results in Section~\ref{sec:main}, and finally, show ablation studies and analyses in Section~\ref{sec:ablation}.
Please see appendix for more results and analyses.

\subsection{Setup}\label{sec:setup}

\mypara{Datasets.}
We use three datasets:
(1) FishNet~\cite{khan2023fishnet}, which contains 17,357 distinct fish species organized taxonomically across five levels.
(2) BIOSCAN-1M~\cite{gharaee2024step}, a collection of high-quality microscope images covering 8,355 insect species.
(3) iNaturalist~\cite{van2018inaturalist}, a dataset spanning 10,000 species of plants and animals, beyond just fish or insects.
All three datasets provide taxonomic labels from \code{Species} to higher ranks like \code{Genus}, \code{Family}, \code{Order}, and \code{Class}. FishNet and BIOSCAN focus on single taxonomic groups, highlighting the challenge of fine-grained generation within the same category due to subtle yet biologically meaningful differences. Evaluations on iNaturalist further demonstrate the generalizability of our method beyond single-species settings. We show results for FishNet and iNaturalist in the main paper, with BIOSCAN-1M in the appendix.

\mypara{Evaluation Metrics.}
We focus on evaluating image quality and text-image alignment.
(1) Image quality: We measure visual fidelity with commonly used Fréchet Inception Distance (FID)~\cite{Seitzer2020FID} and Learned Perceptual Image Patch Similarity (LPIPS)~\cite{zhang2018unreasonable}. 
(2) Taxonomy-image alignment: We use BioCLIP \cite{stevens2024bioclip}, which evaluates how well generated images match hierarchical taxonomy levels. BioCLIP prompts reflect these levels progressively, and higher scores indicate better species-specific alignment.

\mypara{Baselines.} We consider baseline models:
(1) Zero-shot Stable Diffusion (SD)~\cite{rombach2022high}: the vanilla Stable Diffusion model without fine-tuning.
(2) LoRA Fine-tuned Stable Diffusion (SD + LoRA)~\cite{hu2021lora}: fine-tuning Stable Diffusion using Low-Rank Adaptation (LoRA).
(3) Fully fine-tuned Stable Diffusion (SD + Full): fully fine-tuning Stable Diffusion on our dataset, assessing the impact of intensive adaptation on hierarchical image generation. (4) FineDiffusion~\cite{pan2024finediffusion}: state-of-the-art method for fine-grained generation on iNaturalist. To ensure a fair evaluation, we supply hierarchical taxonomy cues (kingdom, phylum, class, order, family, genus, and species) to both the SD + LoRA and + Full baselines.

\mypara{Implementation Details.}
We use the SD v1.5~\cite{rombach2022high} by adding progressive conditioning modules. We add LoRA modules to self- and cross-attention layers. During training, we use AdamW optimizer with a two-stage learning rate: 1 $\times$ 10$^{-4}$ initially (for LoRA and the first taxonomy layer) and 1 $\times$ 10$^{-5}$ for progressive training. We train each level for 250K iterations. During inference, we employ 250 denoising steps with a guidance scale of 6, balancing fidelity and taxonomy alignment. Please see appendix for details.

\begin{table*}[ht]
\small
\tabcolsep 4.5pt
\centering
\caption{\textbf{Quantitative results on FishNet}~\cite{khan2023fishnet}. We report FID and LPIPS to assess image generation quality and BioCLIP for text-to-image alignment, where the text corresponds to the species ``taxonomic name.'' We generate 10 images per category at each level for evaluation, i.e., \code{Order}, \code{Family}, \code{Genus}, and \code{Species}. \ourmethod outperforms the baselines for generating species-specific images.}
\vspace{-3mm}
\begin{tabular}{c|cccccccc|cccc}
\toprule

 \multirow{2.5}{*}{{Method}}    & \multicolumn{4}{c|}{{FID} 
 $\downarrow$}                                                                                          & \multicolumn{4}{c|}{{LPIPS} $\downarrow$}                                                                       & \multicolumn{4}{c}{{BioCLIP - Score} $\uparrow$}                                                                 \\ 
\cmidrule{2-13} 
                             & \multicolumn{1}{c}{Order}   & \multicolumn{1}{c}{Family}   & \multicolumn{1}{c}{Genus}   & \multicolumn{1}{c|}{Species}   & \multicolumn{1}{c}{Order}    & \multicolumn{1}{c}{Family}    & \multicolumn{1}{c}{Genus}    & Species    & \multicolumn{1}{c}{Order}   & \multicolumn{1}{c}{Family}   & \multicolumn{1}{c}{Genus}   & Species   \\ 
\midrule
SD~\cite{rombach2022high}                           & \multicolumn{1}{c}{74.98} & \multicolumn{1}{c}{67.46} & \multicolumn{1}{c}{65.12} & \multicolumn{1}{c|}{61.93} & \multicolumn{1}{c}{0.7854} & \multicolumn{1}{c}{0.7859} & \multicolumn{1}{c}{0.7821} & 0.7737 & \multicolumn{1}{c}{7.48}  & \multicolumn{1}{c}{8.12}  & \multicolumn{1}{c}{8.15}  & 3.35  \\ 
SD + LoRA~\cite{hu2021lora}                    & \multicolumn{1}{c}{56.12} & \multicolumn{1}{c}{52.81} & \multicolumn{1}{c}{45.41} & \multicolumn{1}{c|}{43.91} & \multicolumn{1}{c}{0.7698} & \multicolumn{1}{c}{0.7705} & \multicolumn{1}{c}{0.7617} & 0.7574 & \multicolumn{1}{c}{10.11} & \multicolumn{1}{c}{10.77} & \multicolumn{1}{c}{12.20}  & 7.61  \\ 
SD + Full~\cite{rombach2022high}                    & \multicolumn{1}{c}{50.72} & \multicolumn{1}{c}{47.35} & \multicolumn{1}{c}{41.72} & \multicolumn{1}{c|}{39.41} & \multicolumn{1}{c}{0.7536} & \multicolumn{1}{c}{0.7616} & \multicolumn{1}{c}{0.7582} & 0.7574 & \multicolumn{1}{c}{11.83} & \multicolumn{1}{c}{11.98} & \multicolumn{1}{c}{13.74} & 8.31  \\ 
\midrule
\textbf{\ourmethod (ours)} & \multicolumn{1}{c}{\textbf{41.92}} & \multicolumn{1}{c}{\textbf{40.16}} & \multicolumn{1}{c}{\textbf{27.35}} & \multicolumn{1}{c|}{\textbf{31.87}} & \multicolumn{1}{c}{\textbf{0.7303}} & \multicolumn{1}{c}{\textbf{0.7324}} & \multicolumn{1}{c}{\textbf{0.7349}} & \textbf{0.7319} & \multicolumn{1}{c}{\textbf{15.06}} & \multicolumn{1}{c}{\textbf{16.67}} & \multicolumn{1}{c}{\textbf{18.42}} & \textbf{10.43} \\ 
\bottomrule
\end{tabular}
\vspace{-2mm}
\label{tab:result_vs_baselines}
\end{table*}

\begin{figure*}[t]
    \centering
    \includegraphics[width=0.9\textwidth]{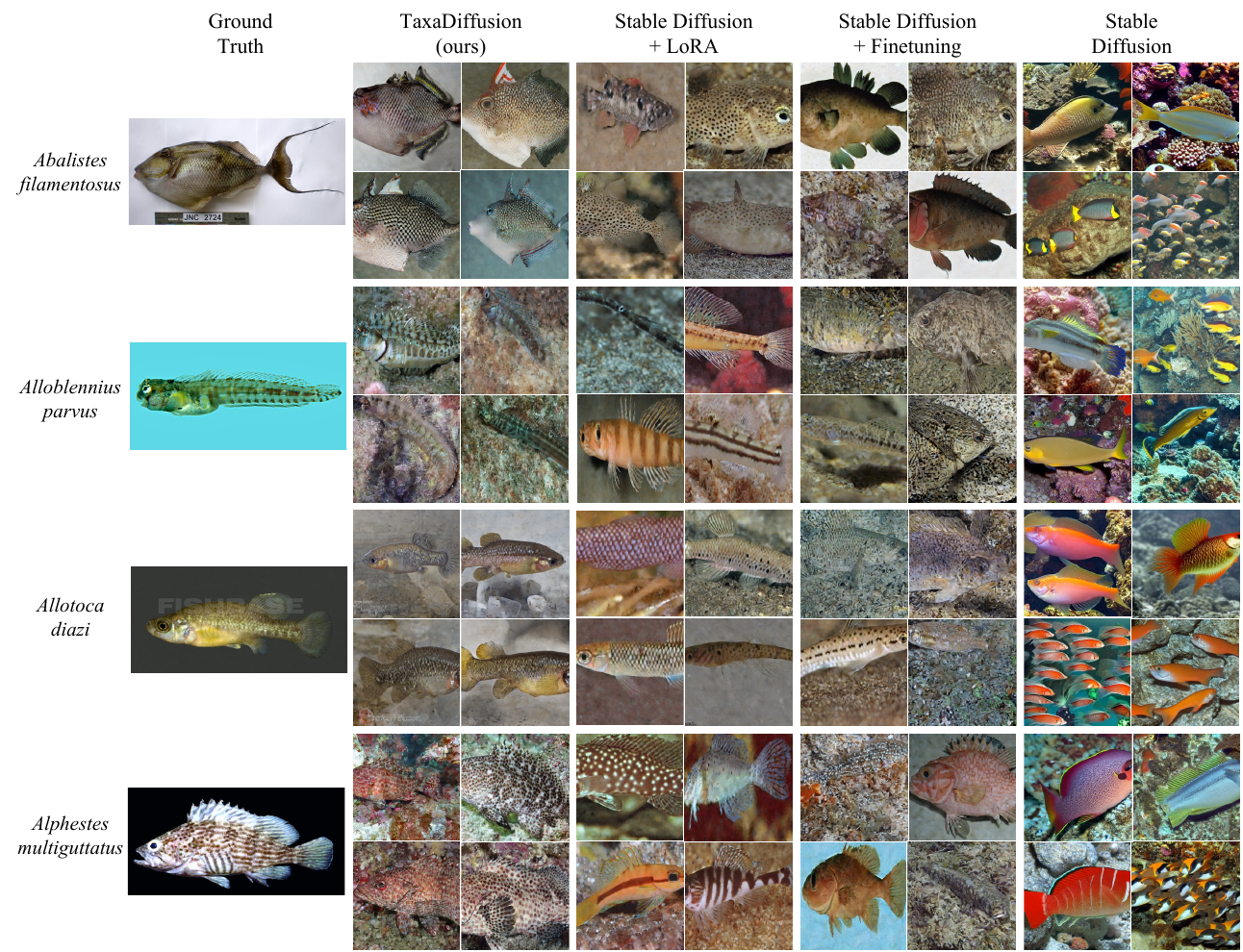}
    \caption{\small \textbf{Qualitative comparison on FishNet}~\cite{khan2023fishnet}.
    We show four samples from each method and observe that \ourmethod captures fine-grained traits (fin structure, body shape, and pattern) more accurately than the baselines which fail to generate the correct species.
    }
    \vspace{-4mm}
    \label{fig:samples}
\end{figure*}

\subsection{Main Results on Fine-Grained Generation}\label{sec:main}
We first evaluate our approach on FishNet~\cite{khan2023fishnet}, a challenging dataset focused on fish species. Fine-grained generation within a species is particularly difficult, as the differences are subtle yet biologically significant. We then extend our evaluation to iNaturalist~\cite{van2018inaturalist}. Please see appendix for additional results on BIOSCAN-1M~\cite{gharaee2024step}.

\mypara{Results on FishNet}~\cite{khan2023fishnet}.
Table~\ref{tab:result_vs_baselines} presents the quantitative results of \ourmethod compared to baseline models across various taxonomic levels. \ourmethod consistently achieves better FID and LPIPS scores, highlighting its capability to generate high-fidelity images that accurately capture level-specific traits. At the \code{Order} level, \ourmethod achieves an FID of 41.9, significantly outperforming all baselines. 
For taxonomy-image alignment, \ourmethod achieves the highest BioCLIP scores across all levels. Figure~\ref{fig:samples} provides a qualitative comparison for four species with limited training examples, alongside their ground truth images. \ourmethod excels at capturing fine-grained features, such as fin structure, body shape, and patterns, resulting in more accurate and biologically correct species representations.

\begin{table*}[t]
\centering
\tabcolsep 4.5pt
\small
\caption{\textbf{Quantitative results for iNaturalist}~\cite{van2018inaturalist}. We report FID and LPIPS to assess image generation quality and use BioCLIP for text-to-image alignment, where the text corresponds to the species ``taxonomic name''. The results highlight \ourmethod outperforms the baseline for generating species-specific images. }
\begin{tabular}{c|cccccccc|cccc}
\toprule
\multirow{2.5}{*}{\textbf{Method}}
  & \multicolumn{4}{c|}{\textbf{FID} $\downarrow$}
  & \multicolumn{4}{c|}{\textbf{LPIPS} $\downarrow$}
  & \multicolumn{4}{c}{\textbf{BioCLIP Score} $\uparrow$} \\ \cmidrule{2-13}
  & Order & Family & Genus & \multicolumn{1}{c|}{Species}
  & Order & Family & Genus & \multicolumn{1}{c|}{Species}
  & Order & Family & Genus & Species \\ \midrule
SD~\cite{rombach2022high}
  & 69.34 & 72.85 & 72.03 & \multicolumn{1}{c|}{73.13}
  & 0.8126 & 0.8102 & 0.8083 & \multicolumn{1}{c|}{0.8134}
  & 7.89   & 9.21   & 7.45   & 6.1 \\ 
\textbf{\ourmethod\ (ours)}
  & \textbf{47.85} & \textbf{49.21} & \textbf{48.94} & \multicolumn{1}{c|}{\textbf{46.39}}
  & \textbf{0.7509} & \textbf{0.7514} & \textbf{0.7501} & \multicolumn{1}{c|}{\textbf{0.7475}}
  & \textbf{9.51}   & \textbf{12.11}  & \textbf{10.30}  & \textbf{10.41} \\ \bottomrule
\end{tabular}
\vspace{-2mm}
\label{tab:iNat_results}
\end{table*}

\begin{figure}[t]
    \centering
    \includegraphics[width=1\linewidth]{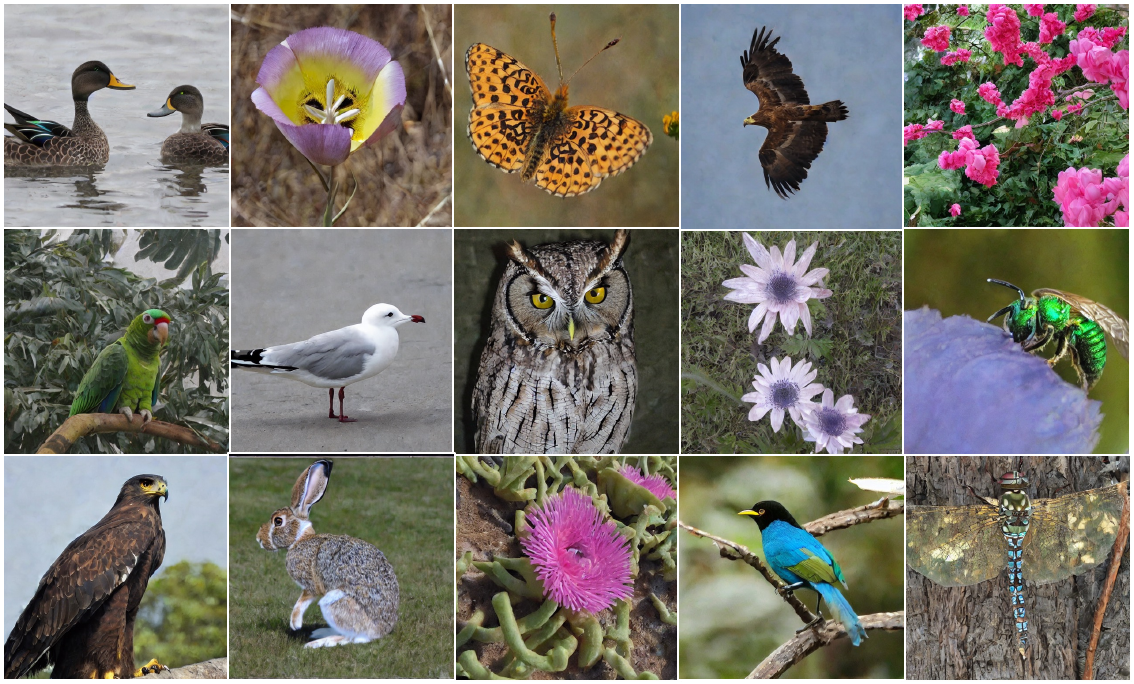}
    \vspace{-4mm}
    \caption{\small \textbf{Qualitative results on iNaturalist} ~\cite{van2018inaturalist}. We show samples from \ourmethod generating morphologically correct traits with diverse background and orientations at the \code{Genus} level. 
    }
    \label{fig:iNat_samples}
\end{figure}

\mypara{Results on iNaturalist}~\cite{van2018inaturalist}.
Beyond single species, we further evaluate on mixed-species on plants and animals using iNaturalist. Table~\ref{tab:iNat_results} provides a quantitative comparison between \ourmethod and the zero-shot Stable Diffusion baseline. The results reveal that the baseline model, when exposed to all taxonomic levels simultaneously, often fail to capture such fine-grained details, thereby highlighting the superiority of our approach in generating biologically consistent images. Figure~\ref{fig:iNat_samples} presents qualitative results of generated images at the \code{Genus} level, highlighting fine-grained visual details across diverse species within the ``Plantae'' and ``Animalia'' kingdoms. \ourmethod, not only preserves high visual fidelity but also accurately captures fine-grained features such as the distinct texture and color patterns, effectively maintaining their morphological traits. Additionally, \ourmethod demonstrates a strong capability to generate samples with varied backgrounds and poses, further underscoring its robustness. 

\begin{table}[t]
\centering
\small
\tabcolsep 7.5pt
\caption{\small \textbf{State-of-the-art comparison.} \ourmethod outperforms FineDiffusion~\cite{pan2024finediffusion}. See appendix for qualitative results.}
\vspace{-2mm}
\begin{tabular}{c|ccc}
\toprule
\multirow{1}{*}{{Method}} & {FID} $\downarrow$ & {LPIPS} $\downarrow$ & {BioCLIP} $\uparrow$ \\ 
\midrule
FineDiffusion~\cite{pan2024finediffusion}   &       74.80               &         0.7613               &   6.46                     \\ 
\textbf{\ourmethod (ours)}         &      \textbf{43.71}             &     \textbf{0.7170}             &   \textbf{8.15}                    \\ 
\bottomrule
\end{tabular}

\label{tab:sota_comparision}
\end{table}

\mypara{Comparison with State-of-the-Art Methods.}
FineDiffusion~\cite{pan2024finediffusion} is a recent approach that aligns closely with our focus on fine-grained image generation. Specifically, it leverages taxonomic information at two levels: ``superclass'' and ``subclass'', using these as conditions to fine-tune a pre-trained DiT-XL/2 model \cite{peebles2023scalable}.  In contrast, \ourmethod advances this by employing fine-grained representations derived from a deeper seven-level taxonomic hierarchy, allowing for a more nuanced understanding of species relationships. Additionally, \ourmethod can generate images of species at all taxonomic levels, facilitating trait discovery by visualizing shared traits at higher taxonomic levels.

FineDiffusion is trained on the iNaturalist dataset \cite{van2018inaturalist}, and Table~\ref{tab:sota_comparision} presents results on 171 species common to both methods. We see that \ourmethod achieves significantly lower FID (43.71 vs. 74.80) and LPIPS (0.7170 vs. 0.7613) compared to FineDiffusion, demonstrating better image generation quality. Furthermore, \ourmethod also achieves a higher BioCLIP score, highlighting better alignment of generated images to the taxonomy. These results demonstrate the effectiveness of \ourmethod, achieving superior performance even when compared to FineDiffusion, which utilizes the DiT-XL/2 model, a framework shown to outperform U-Net-based architectures \cite{peebles2023scalable}.
Qualitative comparisons are illustrated in the Appendix.

\subsection{Ablations and Analyses}\label{sec:ablation}
For the following ablation studies and analyses, we will focus on our approach on the FishNet dataset.

\begin{table}[t]
\centering
\small
\tabcolsep 8.5pt
\caption{ \small \textbf{Importance of progressive training.} \ourmethod achieves competitive results compared with other training strategies. \textbf{All:} train with all taxonomy conditions simultaneously. \textbf{Random:} train with randomly selected taxonomy levels.}
\vspace{-2mm}
\begin{tabular}{c|ccc}
\toprule
{Strategy}  & {FID}   $\downarrow$                        & {LPIPS}  $\downarrow$                        & {BioCLIP}    $\uparrow$         \\ \midrule
All    & 32.43 & 0.7487 & 9.32 \\ 
Random & \textbf{29.53} & 0.7429 & 9.85 \\ 
\midrule
\textbf{Progressive (ours)}    &  31.87 & \textbf{0.7319} &  \textbf{10.43} \\ 
\bottomrule
\end{tabular}
\vspace{-2mm}
\label{tab:progressive:training}
\end{table}

\begin{figure}[t]
    \centering
    \includegraphics[width=1\linewidth]{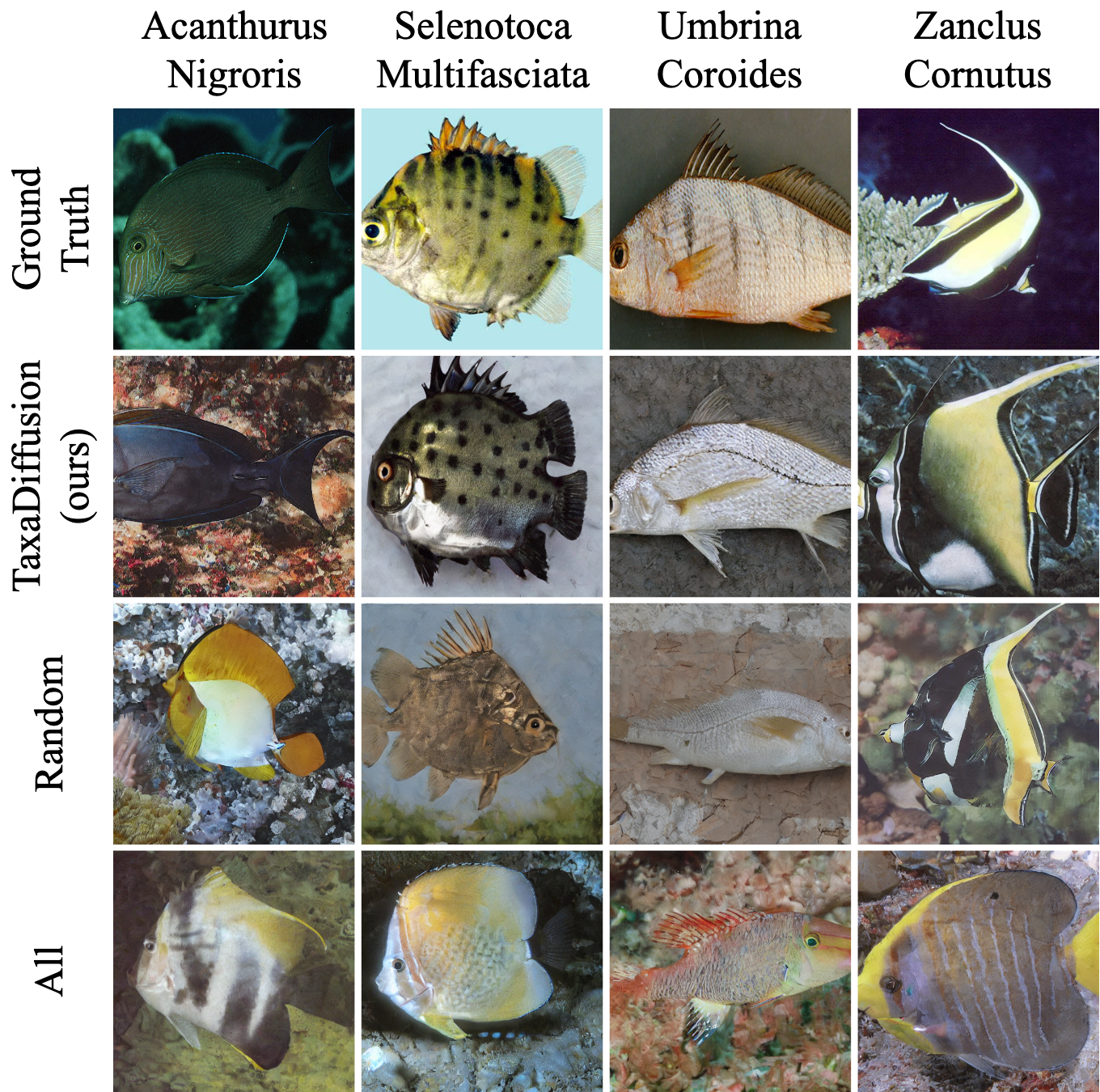}
    \vspace{-5mm}
    \caption{\small \textbf{Ablation of \ourmethod.}  Comparison of generated images using different training strategies in taxonomy-driven conditional diffusion models. The columns represent four species.
    }
    \label{fig:progressive:training}
    \vspace{-2mm}
\end{figure}

\mypara{Importance of Progressive Training.} 
We investigate the impact of different training strategies on the performance of our taxonomy-driven conditional diffusion model. We compare three strategies: (1) \textbf{All}, where the model is trained with all taxonomy conditions simultaneously. (2) \textbf{Random}, where each batch is trained with randomly selected taxonomy levels. (3) \textbf{\ourmethod}, our progressive training approach that sequentially incorporates taxonomy levels from broader to more specific. As shown in Figure~\ref{fig:progressive:training} and Table~\ref{tab:progressive:training}, \ourmethod achieves competitive results. We note that ``random training'' has a slightly lower FID score, but it fails to match the detail and specificity achieved by \ourmethod. The results suggest that our progressive training approach enhances the model’s ability to incorporate taxonomy information effectively, resulting in higher quality and more biologically accurate image generation.

\begin{figure}[t]
    \centering
    \includegraphics[width=1\linewidth]{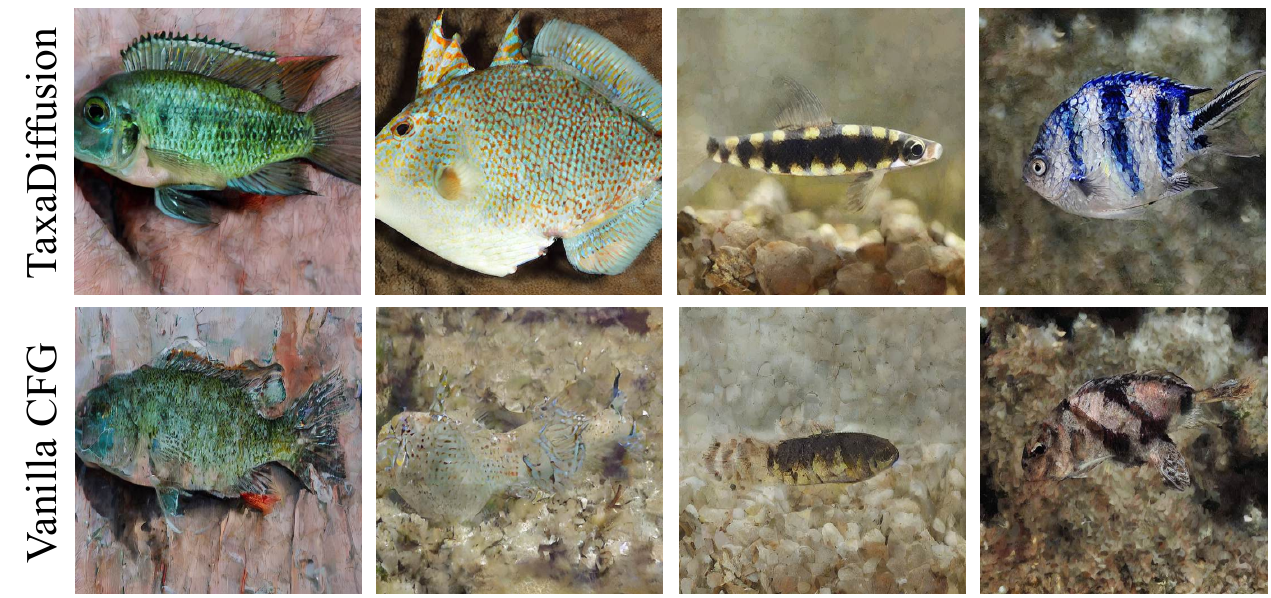}
    \vspace{-5mm}
    \caption{\small \textbf{Ablation of guidance.} Comparison of vanilla CFG with our taxonomy-informed guidance for species generation.
    \label{fig:taxadiff:abla}
    }
    \vspace{-2mm}
\end{figure}

\mypara{Effect of the Condition.}
We conduct an analysis comparing \ourmethod, as described in Section~\ref{sec:guidance}, with the Classifier-Free Guidance (CFG) to assess their influence on fine-grained image generation. In vanilla CFG, conditional and unconditional score estimates are combined as shown in Equation~\ref{eq:cfg}, treating all levels equally and providing a baseline for comparison. In contrast, \ourmethod replaces the unconditional signal with the conditioning from \code{Kingdom} level ($\mathbf{c}^{(0)}$ in Equation~\ref{eq:taxadiffG}), ensuring that guidance starts from the highest taxonomy level and progressively incorporates each level down to the specific traits. This guides the model to capture distinguishing traits effectively, highlighting how hierarchical cues affect the ability the diffusion model to generate morphologically accurate species images. 
Figure~\ref{fig:taxadiff:abla} illustrates that images generated by \ourmethod capture species-specific details more effectively, while vanilla CFG results in less accurate representations. Table~\ref{tab:taxadiff:abla} supports these qualitative findings, with \ourmethod achieving a lower FID (indicating high image quality) and a higher BioCLIP score, highlighting improved biological relevance.

\begin{table}[t]
\centering
\small
\tabcolsep 7.5pt
\caption{\small Ablation study comparing the vanilla CFG with our taxonomy-driven model using FID, LPIPS, and BioCLIP.}
\vspace{-3mm}
\begin{tabular}{c|ccc}
\toprule
{Guidance}  & {FID}  $\downarrow$   & {LPIPS}  $\downarrow$  & {BioCLIP} $\uparrow$ \\ \midrule
Vanilla CFG~\cite{ho2021classifierfree} & 47.61 & 0.7313 & 9.45            \\ 
\textbf{\ourmethod (ours)} & \textbf{32.42}  & \textbf{0.7092}  & \textbf{11.53}            \\ 
\bottomrule
\end{tabular}
\label{tab:taxadiff:abla}
\end{table}

\mypara{Trait Discovery.}
Discovering traits specifically \textit{evolutionary traits} across the taxonomy is crucial for biologists to understand how organisms have diversified over time. Leveraging taxonomy guidance, \ourmethod offers a pathway to capture these evolutionary traits. In Figure \ref{fig:overall},
we see that for \code{Family:} ``Chaetodontidae'' (commonly known as Butterflyfish), have a black stripe or patch through their eyes as a disguise for their heads. Another characteristic trait is the elongated dorsal fin (fin on top of the fish) that sets ``Chaetodontidae'' apart from other families ``Acanthuridae'' and ``Zanclidae'' in the same \code{Order:} ``Acanthuriformers''. 
We observe the emergence of new traits as we go from \code{Family} to \code{Genus} level. Within \code{Genus:} ``Amphichaetodon'', black vertical stripes are prominent, while \code{Genus:} ``Chaetodon'' exhibits a large black patch near the caudal (tail) fin. \ourmethod retains shared traits (black stripe around eyes) while acquiring genus-specific traits (body strip patterns). 

\section{Conclusion and Discussion}
\label{sec:disc}

In this paper, we propose \ourmethod, a novel approach for leveraging hierarchical taxonomy to enhance fine-grained species generation in diffusion models. Our key novelty lies in the effective use of taxonomy specifically for model training. By progressively training the model through different taxonomic levels, we enable the model to capture both broad characteristics and subtle distinctions among species. \ourmethodG allows conditioning at any taxonomy level during inference, guiding the model to focus on species-specific traits by highlighting the differences between a species and its ancestor nodes. Our method effectively teaches the diffusion model to understand level-specific knowledge encoded in the taxonomy tree, resulting in high-fidelity, morphologically accurate images that reflect species identity.  We also show, that for rare species with limited samples (e.g., ``Amphichaetodon Melbae'' with $1$ sample), taxonomic guidance enables effective knowledge transfer from sibling species, thereby generating accurate \& morphologically rich images.
Experimental results demonstrate that \ourmethod outperforms baseline models to produce images that are not only visually convincing but also semantically aligned with biological taxonomy.

\mypara{Limitation and Future Work.} 
While \ourmethod significantly improves fine-grained, object-centric species generation, it has limitations to address in future work. Our current U-Net architecture, though effective, may not fully leverage scalable transformer-based models like Diffusion Transformers \cite{peebles2023scalable}. Incorporating transformers could enhance scalability and performance.

\section{Acknowledgments}
This research is supported in part by grants from the National Science Foundation (OAC-2118240, HDR Institute: Imageomics).

{
    \small
    \bibliographystyle{ieeenat_fullname}
    \bibliography{main}

\begin{thebibliography}{61}
\providecommand{\natexlab}[1]{#1}
\providecommand{\url}[1]{\texttt{#1}}
\expandafter\ifx\csname urlstyle\endcsname\relax
  \providecommand{\doi}[1]{doi: #1}\else
  \providecommand{\doi}{doi: \begingroup \urlstyle{rm}\Url}\fi

\bibitem[IDi(2020)]{IDigBio}
idigbio.
\newblock \emph{http://www.idigbio.org/portal}, 2020.

\bibitem[inh(2022)]{inhs}
Inhs collections data.
\newblock \emph{http://biocoll.inhs.illinois.edu/portal/index.php}, 2022.

\bibitem[Chan et~al.(2023)Chan, Nagano, Chan, Bergman, Park, Levy, Aittala, De~Mello, Karras, and Wetzstein]{chan2023generative}
Eric~R Chan, Koki Nagano, Matthew~A Chan, Alexander~W Bergman, Jeong~Joon Park, Axel Levy, Miika Aittala, Shalini De~Mello, Tero Karras, and Gordon Wetzstein.
\newblock Generative novel view synthesis with 3d-aware diffusion models.
\newblock In \emph{Proceedings of the IEEE/CVF International Conference on Computer Vision}, pages 4217--4229, 2023.

\bibitem[Chen et~al.(2024)Chen, Zhang, Liu, Gu, Kong, Yuan, et~al.]{chen2024hierarchical}
Zheng Chen, Yulun Zhang, Ding Liu, Jinjin Gu, Linghe Kong, Xin Yuan, et~al.
\newblock Hierarchical integration diffusion model for realistic image deblurring.
\newblock \emph{Advances in neural information processing systems}, 36, 2024.

\bibitem[Child(2020)]{child2020very}
Rewon Child.
\newblock Very deep vaes generalize autoregressive models and can outperform them on images.
\newblock \emph{arXiv preprint arXiv:2011.10650}, 2020.

\bibitem[Dhariwal and Nichol(2021)]{dhariwal2021diffusion}
Prafulla Dhariwal and Alexander Nichol.
\newblock Diffusion models beat gans on image synthesis.
\newblock \emph{Advances in neural information processing systems}, 34:\penalty0 8780--8794, 2021.

\bibitem[Dubois(2007)]{dubois2007phylogeny}
Alain Dubois.
\newblock Phylogeny, taxonomy and nomenclature: The problem of taxonomic categories and of nomenclatural ranks.
\newblock \emph{Zootaxa}, 1519:\penalty0 27--68, 2007.

\bibitem[Gafni et~al.(2022)Gafni, Polyak, Ashual, Sheynin, Parikh, and Taigman]{gafni2022make}
Oran Gafni, Adam Polyak, Oron Ashual, Shelly Sheynin, Devi Parikh, and Yaniv Taigman.
\newblock Make-a-scene: Scene-based text-to-image generation with human priors.
\newblock In \emph{European Conference on Computer Vision}, pages 89--106. Springer, 2022.

\bibitem[Gharaee et~al.(2024)Gharaee, Gong, Pellegrino, Zarubiieva, Haurum, Lowe, McKeown, Ho, McLeod, Wei, et~al.]{gharaee2024step}
Zahra Gharaee, ZeMing Gong, Nicholas Pellegrino, Iuliia Zarubiieva, Joakim~Bruslund Haurum, Scott Lowe, Jaclyn McKeown, Chris Ho, Joschka McLeod, Yi-Yun Wei, et~al.
\newblock A step towards worldwide biodiversity assessment: The bioscan-1m insect dataset.
\newblock \emph{Advances in Neural Information Processing Systems}, 36, 2024.

\bibitem[Goodfellow et~al.(2014)Goodfellow, Pouget-Abadie, Mirza, Xu, Warde-Farley, Ozair, Courville, and Bengio]{goodfellow2014generative}
Ian Goodfellow, Jean Pouget-Abadie, Mehdi Mirza, Bing Xu, David Warde-Farley, Sherjil Ozair, Aaron Courville, and Yoshua Bengio.
\newblock Generative adversarial nets.
\newblock \emph{Advances in neural information processing systems}, 27, 2014.

\bibitem[Han et~al.(2023)Han, Yang, Kwon, and Ye]{han2023highly}
Inhwa Han, Serin Yang, Taesung Kwon, and Jong~Chul Ye.
\newblock Highly personalized text embedding for image manipulation by stable diffusion.
\newblock \emph{arXiv preprint arXiv:2303.08767}, 2023.

\bibitem[Ho and Salimans(2021)]{ho2021classifierfree}
Jonathan Ho and Tim Salimans.
\newblock Classifier-free diffusion guidance.
\newblock In \emph{NeurIPS 2021 Workshop on Deep Generative Models and Downstream Applications}, 2021.

\bibitem[Ho and Salimans(2022)]{ho2022classifier}
Jonathan Ho and Tim Salimans.
\newblock Classifier-free diffusion guidance.
\newblock \emph{arXiv preprint arXiv:2207.12598}, 2022.

\bibitem[Ho et~al.(2020)Ho, Jain, and Abbeel]{ho2020denoising}
Jonathan Ho, Ajay Jain, and Pieter Abbeel.
\newblock Denoising diffusion probabilistic models.
\newblock \emph{Advances in neural information processing systems}, 33:\penalty0 6840--6851, 2020.

\bibitem[Hobern et~al.(2021)Hobern, Barik, Christidis, T.Garnett, Kirk, Orrell, Pape, Pyle, Thiele, Zachos, and B{\'a}nki]{Hobern2021}
Donald Hobern, Saroj~K. Barik, Les Christidis, Stephen T.Garnett, Paul Kirk, Thomas~M. Orrell, Thomas Pape, Richard~L. Pyle, Kevin~R. Thiele, Frank~E. Zachos, and Olaf B{\'a}nki.
\newblock Towards a global list of accepted species vi: The catalogue of life checklist.
\newblock \emph{Organisms Diversity {\&} Evolution}, 21\penalty0 (4):\penalty0 677--690, 2021.

\bibitem[Hu et~al.(2021)Hu, Shen, Wallis, Allen-Zhu, Li, Wang, Wang, and Chen]{hu2021lora}
Edward~J Hu, Yelong Shen, Phillip Wallis, Zeyuan Allen-Zhu, Yuanzhi Li, Shean Wang, Lu Wang, and Weizhu Chen.
\newblock Lora: Low-rank adaptation of large language models.
\newblock \emph{arXiv preprint arXiv:2106.09685}, 2021.

\bibitem[Huang et~al.(2024)Huang, Huang, Liu, Yan, Lv, Liu, Xiong, Zhang, Chen, and Cao]{huang2024diffusion}
Yi Huang, Jiancheng Huang, Yifan Liu, Mingfu Yan, Jiaxi Lv, Jianzhuang Liu, Wei Xiong, He Zhang, Shifeng Chen, and Liangliang Cao.
\newblock Diffusion model-based image editing: A survey.
\newblock \emph{arXiv preprint arXiv:2402.17525}, 2024.

\bibitem[Karras et~al.(2024)Karras, Aittala, Kynk{\"a}{\"a}nniemi, Lehtinen, Aila, and Laine]{karras2024guiding}
Tero Karras, Miika Aittala, Tuomas Kynk{\"a}{\"a}nniemi, Jaakko Lehtinen, Timo Aila, and Samuli Laine.
\newblock Guiding a diffusion model with a bad version of itself.
\newblock \emph{arXiv preprint arXiv:2406.02507}, 2024.

\bibitem[Khan et~al.(2023)Khan, Li, Temple, and Elhoseiny]{khan2023fishnet}
Faizan~Farooq Khan, Xiang Li, Andrew~J Temple, and Mohamed Elhoseiny.
\newblock Fishnet: A large-scale dataset and benchmark for fish recognition, detection, and functional trait prediction.
\newblock In \emph{Proceedings of the IEEE/CVF International Conference on Computer Vision}, pages 20496--20506, 2023.

\bibitem[Khurana et~al.(2024)Khurana, Daw, Maruf, Uyeda, Dahdul, Charpentier, Bak{\i}{\c{s}}, Bart~Jr, Mabee, Lapp, et~al.]{khurana2024hierarchical}
Mridul Khurana, Arka Daw, M Maruf, Josef~C Uyeda, Wasila Dahdul, Caleb Charpentier, Yasin Bak{\i}{\c{s}}, Henry~L Bart~Jr, Paula~M Mabee, Hilmar Lapp, et~al.
\newblock Hierarchical conditioning of diffusion models using tree-of-life for studying species evolution.
\newblock 2024.

\bibitem[Kim et~al.(2022)Kim, Kwon, and Ye]{kim2022diffusionclip}
Gwanghyun Kim, Taesung Kwon, and Jong~Chul Ye.
\newblock Diffusionclip: Text-guided diffusion models for robust image manipulation.
\newblock In \emph{Proceedings of the IEEE/CVF conference on computer vision and pattern recognition}, pages 2426--2435, 2022.

\bibitem[Kingma(2013)]{kingma2013auto}
Diederik~P Kingma.
\newblock Auto-encoding variational bayes.
\newblock \emph{arXiv preprint arXiv:1312.6114}, 2013.

\bibitem[Li et~al.(2025)Li, Yang, Kuang, Wu, Wang, Xiao, and Chen]{li2025controlnet}
Ming Li, Taojiannan Yang, Huafeng Kuang, Jie Wu, Zhaoning Wang, Xuefeng Xiao, and Chen Chen.
\newblock Controlnet++: Improving conditional controls with efficient consistency feedback.
\newblock In \emph{European Conference on Computer Vision}, pages 129--147. Springer, 2025.

\bibitem[Li et~al.(2023)Li, Liu, Huang, Xia, Yang, and Lu]{li2023progressive}
Peipei Li, Xiyan Liu, Jizhou Huang, Deguo Xia, Jianzhong Yang, and Zhen Lu.
\newblock Progressive generation of 3d point clouds with hierarchical consistency.
\newblock \emph{Pattern Recognition}, 136:\penalty0 109200, 2023.

\bibitem[Li et~al.(2024)Li, Hou, and Loy]{li2024stylegan}
Xiaoming Li, Xinyu Hou, and Chen~Change Loy.
\newblock When stylegan meets stable diffusion: a w+ adapter for personalized image generation.
\newblock In \emph{Proceedings of the IEEE/CVF Conference on Computer Vision and Pattern Recognition}, pages 2187--2196, 2024.

\bibitem[Matsunaga et~al.(2022)Matsunaga, Ishii, Hayakawa, Suzuki, and Narihira]{matsunaga2022fine}
Naoki Matsunaga, Masato Ishii, Akio Hayakawa, Kenji Suzuki, and Takuya Narihira.
\newblock Fine-grained image editing by pixel-wise guidance using diffusion models.
\newblock \emph{arXiv preprint arXiv:2212.02024}, 2022.

\bibitem[Monsefi et~al.(2024)Monsefi, Sailaja, Alilooee, Lim, and Ramnath]{monsefi2024detailclip}
Amin~Karimi Monsefi, Kishore~Prakash Sailaja, Ali Alilooee, Ser-Nam Lim, and Rajiv Ramnath.
\newblock Detailclip: Detail-oriented clip for fine-grained tasks.
\newblock \emph{arXiv preprint arXiv:2409.06809}, 2024.

\bibitem[Navard et~al.(2024)Navard, Monsefi, Zhou, Chao, Yilmaz, and Ramnath]{navard2024knobgen}
Pouyan Navard, Amin~Karimi Monsefi, Mengxi Zhou, Wei-Lun Chao, Alper Yilmaz, and Rajiv Ramnath.
\newblock Knobgen: Controlling the sophistication of artwork in sketch-based diffusion models.
\newblock \emph{arXiv preprint arXiv:2410.01595}, 2024.

\bibitem[Nichol et~al.(2021)Nichol, Dhariwal, Ramesh, Shyam, Mishkin, McGrew, Sutskever, and Chen]{nichol2021glide}
Alex Nichol, Prafulla Dhariwal, Aditya Ramesh, Pranav Shyam, Pamela Mishkin, Bob McGrew, Ilya Sutskever, and Mark Chen.
\newblock Glide: Towards photorealistic image generation and editing with text-guided diffusion models.
\newblock \emph{arXiv preprint arXiv:2112.10741}, 2021.

\bibitem[Padial et~al.(2010)Padial, Miralles, De~la Riva, and Vences]{padial2010integrative}
Jos{\'e}~M Padial, Aur{\'e}lien Miralles, Ignacio De~la Riva, and Miguel Vences.
\newblock The integrative future of taxonomy.
\newblock \emph{Frontiers in zoology}, 7:\penalty0 1--14, 2010.

\bibitem[Pan et~al.(2024)Pan, Wang, Li, He, and Lai]{pan2024finediffusion}
Ziying Pan, Kun Wang, Gang Li, Feihong He, and Yongxuan Lai.
\newblock Finediffusion: Scaling up diffusion models for fine-grained image generation with 10,000 classes.
\newblock \emph{arXiv preprint arXiv:2402.18331}, 2024.

\bibitem[Parihar et~al.(2025)Parihar, Sachidanand, Mani, Karmali, and Venkatesh~Babu]{parihar2025precisecontrol}
Rishubh Parihar, VS Sachidanand, Sabariswaran Mani, Tejan Karmali, and R Venkatesh~Babu.
\newblock Precisecontrol: Enhancing text-to-image diffusion models with fine-grained attribute control.
\newblock In \emph{European Conference on Computer Vision}, pages 469--487. Springer, 2025.

\bibitem[Peebles and Xie(2023)]{peebles2023scalable}
William Peebles and Saining Xie.
\newblock Scalable diffusion models with transformers.
\newblock In \emph{Proceedings of the IEEE/CVF International Conference on Computer Vision}, pages 4195--4205, 2023.

\bibitem[Pi et~al.(2023)Pi, Peng, Yang, Zhou, and Bao]{pi2023hierarchical}
Huaijin Pi, Sida Peng, Minghui Yang, Xiaowei Zhou, and Hujun Bao.
\newblock Hierarchical generation of human-object interactions with diffusion probabilistic models.
\newblock In \emph{Proceedings of the IEEE/CVF International Conference on Computer Vision}, pages 15061--15073, 2023.

\bibitem[Podell et~al.(2023)Podell, English, Lacey, Blattmann, Dockhorn, M{\"u}ller, Penna, and Rombach]{podell2023sdxl}
Dustin Podell, Zion English, Kyle Lacey, Andreas Blattmann, Tim Dockhorn, Jonas M{\"u}ller, Joe Penna, and Robin Rombach.
\newblock Sdxl: Improving latent diffusion models for high-resolution image synthesis.
\newblock \emph{arXiv preprint arXiv:2307.01952}, 2023.

\bibitem[Radford et~al.(2021)Radford, Kim, Hallacy, Ramesh, Goh, Agarwal, Sastry, Askell, Mishkin, Clark, et~al.]{radford2021learning}
Alec Radford, Jong~Wook Kim, Chris Hallacy, Aditya Ramesh, Gabriel Goh, Sandhini Agarwal, Girish Sastry, Amanda Askell, Pamela Mishkin, Jack Clark, et~al.
\newblock Learning transferable visual models from natural language supervision.
\newblock In \emph{International conference on machine learning}, pages 8748--8763. PMLR, 2021.

\bibitem[Ramesh et~al.(2022)Ramesh, Dhariwal, Nichol, Chu, and Chen]{ramesh2022hierarchical}
Aditya Ramesh, Prafulla Dhariwal, Alex Nichol, Casey Chu, and Mark Chen.
\newblock Hierarchical text-conditional image generation with clip latents.
\newblock \emph{arXiv preprint arXiv:2204.06125}, 1\penalty0 (2):\penalty0 3, 2022.

\bibitem[Razavi et~al.(2019)Razavi, Van~den Oord, and Vinyals]{razavi2019generating}
Ali Razavi, Aaron Van~den Oord, and Oriol Vinyals.
\newblock Generating diverse high-fidelity images with vq-vae-2.
\newblock \emph{Advances in neural information processing systems}, 32, 2019.

\bibitem[Rombach et~al.(2022)Rombach, Blattmann, Lorenz, Esser, and Ommer]{rombach2022high}
Robin Rombach, Andreas Blattmann, Dominik Lorenz, Patrick Esser, and Bj{\"o}rn Ommer.
\newblock High-resolution image synthesis with latent diffusion models.
\newblock In \emph{Proceedings of the IEEE/CVF conference on computer vision and pattern recognition}, pages 10684--10695, 2022.

\bibitem[Ruiz et~al.(2023)Ruiz, Li, Jampani, Pritch, Rubinstein, and Aberman]{ruiz2023dreambooth}
Nataniel Ruiz, Yuanzhen Li, Varun Jampani, Yael Pritch, Michael Rubinstein, and Kfir Aberman.
\newblock Dreambooth: Fine tuning text-to-image diffusion models for subject-driven generation.
\newblock In \emph{Proceedings of the IEEE/CVF conference on computer vision and pattern recognition}, pages 22500--22510, 2023.

\bibitem[Ruiz et~al.(2024)Ruiz, Li, Jampani, Wei, Hou, Pritch, Wadhwa, Rubinstein, and Aberman]{ruiz2024hyperdreambooth}
Nataniel Ruiz, Yuanzhen Li, Varun Jampani, Wei Wei, Tingbo Hou, Yael Pritch, Neal Wadhwa, Michael Rubinstein, and Kfir Aberman.
\newblock Hyperdreambooth: Hypernetworks for fast personalization of text-to-image models.
\newblock In \emph{Proceedings of the IEEE/CVF Conference on Computer Vision and Pattern Recognition}, pages 6527--6536, 2024.

\bibitem[Saharia et~al.(2022)Saharia, Chan, Saxena, Li, Whang, Denton, Ghasemipour, Gontijo~Lopes, Karagol~Ayan, Salimans, et~al.]{saharia2022photorealistic}
Chitwan Saharia, William Chan, Saurabh Saxena, Lala Li, Jay Whang, Emily~L Denton, Kamyar Ghasemipour, Raphael Gontijo~Lopes, Burcu Karagol~Ayan, Tim Salimans, et~al.
\newblock Photorealistic text-to-image diffusion models with deep language understanding.
\newblock \emph{Advances in neural information processing systems}, 35:\penalty0 36479--36494, 2022.

\bibitem[Schuhmann et~al.(2022)Schuhmann, Beaumont, Vencu, Gordon, Wightman, Cherti, Coombes, Katta, Mullis, Wortsman, et~al.]{schuhmann2022laion}
Christoph Schuhmann, Romain Beaumont, Richard Vencu, Cade Gordon, Ross Wightman, Mehdi Cherti, Theo Coombes, Aarush Katta, Clayton Mullis, Mitchell Wortsman, et~al.
\newblock Laion-5b: An open large-scale dataset for training next generation image-text models.
\newblock \emph{Advances in Neural Information Processing Systems}, 35:\penalty0 25278--25294, 2022.

\bibitem[Seitzer(2020)]{Seitzer2020FID}
Maximilian Seitzer.
\newblock {pytorch-fid: FID Score for PyTorch}.
\newblock \url{https://github.com/mseitzer/pytorch-fid}, 2020.
\newblock Version 0.3.0.

\bibitem[Stevens et~al.(2024)Stevens, Wu, Thompson, Campolongo, Song, Carlyn, Dong, Dahdul, Stewart, Berger-Wolf, et~al.]{stevens2024bioclip}
Samuel Stevens, Jiaman Wu, Matthew~J Thompson, Elizabeth~G Campolongo, Chan~Hee Song, David~Edward Carlyn, Li Dong, Wasila~M Dahdul, Charles Stewart, Tanya Berger-Wolf, et~al.
\newblock Bioclip: A vision foundation model for the tree of life.
\newblock In \emph{Proceedings of the IEEE/CVF Conference on Computer Vision and Pattern Recognition}, pages 19412--19424, 2024.

\bibitem[Tan et~al.(2024)Tan, Yang, Qin, Yang, Qian, Zhou, Zhang, and Li]{tan2024empirical}
Zhiyu Tan, Mengping Yang, Luozheng Qin, Hao Yang, Ye Qian, Qiang Zhou, Cheng Zhang, and Hao Li.
\newblock An empirical study and analysis of text-to-image generation using large language model-powered textual representation.
\newblock In \emph{European Conference on Computer Vision}, pages 472--489. Springer, 2024.

\bibitem[Vahdat and Kautz(2020)]{vahdat2020nvae}
Arash Vahdat and Jan Kautz.
\newblock Nvae: A deep hierarchical variational autoencoder.
\newblock \emph{Advances in neural information processing systems}, 33:\penalty0 19667--19679, 2020.

\bibitem[Van~Horn et~al.(2018)Van~Horn, Mac~Aodha, Song, Cui, Sun, Shepard, Adam, Perona, and Belongie]{van2018inaturalist}
Grant Van~Horn, Oisin Mac~Aodha, Yang Song, Yin Cui, Chen Sun, Alex Shepard, Hartwig Adam, Pietro Perona, and Serge Belongie.
\newblock The inaturalist species classification and detection dataset.
\newblock In \emph{Proceedings of the IEEE conference on computer vision and pattern recognition}, pages 8769--8778, 2018.

\bibitem[Wang et~al.(2024{\natexlab{a}})Wang, Huang, Shi, Bian, Song, Liu, and Li]{wang2024animatelcm}
Fu-Yun Wang, Zhaoyang Huang, Xiaoyu Shi, Weikang Bian, Guanglu Song, Yu Liu, and Hongsheng Li.
\newblock Animatelcm: Accelerating the animation of personalized diffusion models and adapters with decoupled consistency learning.
\newblock \emph{arXiv preprint arXiv:2402.00769}, 2024{\natexlab{a}}.

\bibitem[Wang et~al.(2024{\natexlab{b}})Wang, Sun, Tan, Chen, Chen, Li, Zhang, and Song]{wang2024towards}
Junyan Wang, Zhenhong Sun, Zhiyu Tan, Xuanbai Chen, Weihua Chen, Hao Li, Cheng Zhang, and Yang Song.
\newblock Towards effective usage of human-centric priors in diffusion models for text-based human image generation.
\newblock In \emph{Proceedings of the IEEE/CVF Conference on Computer Vision and Pattern Recognition}, pages 8446--8455, 2024{\natexlab{b}}.

\bibitem[Wang et~al.(2023)Wang, Leng, Li, Wu, and Liang]{wang2023fg}
Yin Wang, Zhiying Leng, Frederick~WB Li, Shun-Cheng Wu, and Xiaohui Liang.
\newblock Fg-t2m: Fine-grained text-driven human motion generation via diffusion model.
\newblock In \emph{Proceedings of the IEEE/CVF International Conference on Computer Vision}, pages 22035--22044, 2023.

\bibitem[Xu et~al.(2021)Xu, Shen, Zhu, Yang, and Zhou]{xu2021generative}
Yinghao Xu, Yujun Shen, Jiapeng Zhu, Ceyuan Yang, and Bolei Zhou.
\newblock Generative hierarchical features from synthesizing images.
\newblock In \emph{Proceedings of the IEEE/CVF Conference on Computer Vision and Pattern Recognition}, pages 4432--4442, 2021.

\bibitem[Ye et~al.(2023)Ye, Li, Zhou, Jiale, Yu, Luo, Song, Xing, Zhang, and Yang]{ye2023fine}
Yuteng Ye, Guanwen Li, Hang Zhou, Cai Jiale, Junqing Yu, Yawei Luo, Zikai Song, Qilong Xing, Youjia Zhang, and Wei Yang.
\newblock Fine-grained appearance transfer with diffusion models.
\newblock \emph{arXiv preprint arXiv:2311.16513}, 2023.

\bibitem[Yue et~al.(2023)Yue, Guo, Ning, Cui, Zhu, and Yuan]{yue2023chatface}
Dongxu Yue, Qin Guo, Munan Ning, Jiaxi Cui, Yuesheng Zhu, and Li Yuan.
\newblock Chatface: Chat-guided real face editing via diffusion latent space manipulation.
\newblock \emph{arXiv preprint arXiv:2305.14742}, 2023.

\bibitem[Zeng et~al.(2024)Zeng, Dong, Peers, Kong, Wu, and Tong]{zeng2024dilightnet}
Chong Zeng, Yue Dong, Pieter Peers, Youkang Kong, Hongzhi Wu, and Xin Tong.
\newblock Dilightnet: Fine-grained lighting control for diffusion-based image generation.
\newblock In \emph{ACM SIGGRAPH 2024 Conference Papers}, pages 1--12, 2024.

\bibitem[Zhang et~al.(2023{\natexlab{a}})Zhang, Chen, Chai, Wu, Lagun, Beeler, and De~la Torre]{zhang2023iti}
Cheng Zhang, Xuanbai Chen, Siqi Chai, Chen~Henry Wu, Dmitry Lagun, Thabo Beeler, and Fernando De~la Torre.
\newblock Iti-gen: Inclusive text-to-image generation.
\newblock In \emph{Proceedings of the IEEE/CVF International Conference on Computer Vision}, pages 3969--3980, 2023{\natexlab{a}}.

\bibitem[Zhang et~al.(2023{\natexlab{b}})Zhang, Rao, and Agrawala]{zhang2023adding}
Lvmin Zhang, Anyi Rao, and Maneesh Agrawala.
\newblock Adding conditional control to text-to-image diffusion models.
\newblock In \emph{Proceedings of the IEEE/CVF International Conference on Computer Vision}, pages 3836--3847, 2023{\natexlab{b}}.

\bibitem[Zhang et~al.(2018)Zhang, Isola, Efros, Shechtman, and Wang]{zhang2018unreasonable}
Richard Zhang, Phillip Isola, Alexei~A Efros, Eli Shechtman, and Oliver Wang.
\newblock The unreasonable effectiveness of deep features as a perceptual metric.
\newblock In \emph{Proceedings of the IEEE conference on computer vision and pattern recognition}, pages 586--595, 2018.

\bibitem[Zhang et~al.(2024)Zhang, Wei, Zhang, Wu, Zhang, Lei, and Li]{zhang2024survey}
Xulu Zhang, Xiao-Yong Wei, Wengyu Zhang, Jinlin Wu, Zhaoxiang Zhang, Zhen Lei, and Qing Li.
\newblock A survey on personalized content synthesis with diffusion models.
\newblock \emph{arXiv preprint arXiv:2405.05538}, 2024.

\bibitem[Zhang et~al.(2023{\natexlab{c}})Zhang, Han, Ghosh, Metaxas, and Ren]{zhang2023sine}
Zhixing Zhang, Ligong Han, Arnab Ghosh, Dimitris~N Metaxas, and Jian Ren.
\newblock Sine: Single image editing with text-to-image diffusion models.
\newblock In \emph{Proceedings of the IEEE/CVF Conference on Computer Vision and Pattern Recognition}, pages 6027--6037, 2023{\natexlab{c}}.

\bibitem[Zhao et~al.(2024)Zhao, Chen, Chen, Bao, Hao, Yuan, and Wong]{zhao2024uni}
Shihao Zhao, Dongdong Chen, Yen-Chun Chen, Jianmin Bao, Shaozhe Hao, Lu Yuan, and Kwan-Yee~K Wong.
\newblock Uni-controlnet: All-in-one control to text-to-image diffusion models.
\newblock \emph{Advances in Neural Information Processing Systems}, 36, 2024.

\end{thebibliography}
}
\clearpage
\maketitlesupplementary

\setcounter{figure}{0}
\setcounter{table}{0}
\renewcommand\thesection{\Alph{section}}
\renewcommand{\thetable}{S\arabic{table}}  
\renewcommand{\thefigure}{S\arabic{figure}}

\appendix

\begin{center}
{
    \hypersetup{linkcolor=citecolor,linktocpage=true}
    \tableofcontents
}
\end{center}

\gradientline

\section{Additional Implementation Details}

We provide implementation details omitted in the main text. Our model and code will be publicly available.

\subsection{Hyperparameters}

Table~\ref{tab:hyperparameter} summarizes the hyperparameters used across the models, including the baselines and our proposed \ourmethod. All methods use the Stable Diffusion v1.5 \cite{rombach2022high} as the base generator. We employ consistent settings, such as the image resolution (\(512 \times 512\)) and latent space dimensions (\textit{z}-shape: \(64 \times 64 \times 4\)), to ensure comparability.

Unlike full fine-tuning, which requires updating all \(859.52\) million parameters of the model, our method, \ourmethod, trains only a subset of parameters --- accounting for just \(6.5\%\) of the total model size. This approach, utilizing \(55.93\) million trainable parameters, represents a significantly smaller fraction compared to full fine-tuning. Despite this reduction, our progressive taxonomy --- based training strategy amplifies the impact of these updates by leveraging hierarchical information from higher taxonomic levels. This enables our method to achieve superior generative performance with a considerably smaller training footprint, highlighting the efficiency and effectiveness of our targeted approach.

We use a guidance scale of \(6\) across all models during inference, balancing image fidelity and trait specificity (see Section \ref{sup:sec:taxaguide:scale} for more details). 
We set diffusion sampling step to $250$ for consistent evaluation across all approaches.

We use cross-attention for all models as the conditioning mechanism and a batch size of $12$ per GPU and a total batch size of \(192\).
We use a learning rate of $1\times10^{-4}$ for \textit{SD + LoRA} and $1\times10^{-5}$ for \textit{SD + Full Finetuning}.
For our method \ourmethod, we use a learning rate of $1\times10^{-4}$ for the first step because of having LoRA, and then $1\times10^{-5}$ for the rest of the steps.

\begin{table*}
\tabcolsep 18pt 
\centering
\small
\centering\caption{\textbf{Hyperparameter setting}. Training and inference stage hyperparameter details of all the baselines and \ourmethod. $\dagger$: We use a learning rate of $1\times10^{-4}$ only for training the first level of \ourmethod and the LoRA stage. *: \ourmethod steps are mentioned for each level training. SD: Stable Diffusion 1.5.}
\begin{tabular}{l|cccc}
\toprule
      
Parameter   & SD &    SD + LoRA & SD + Full Finetuning & \ourmethod \\   
\midrule
    Image resolution    & 512 $\times$ 512 & 512 $\times$ 512 & 512 $\times$ 512 & 512 $\times$ 512 \\
    \textit{z}-shape    &  64 $\times$ 64 $\times$ 4 &     64 $\times$ 64 $\times$ 4 &  64 $\times$ 64 $\times$ 4  & 64 $\times$ 64 $\times$ 4 \\
    Model Size    &  859.52 M & 860.32 M  & 859.52 M & 915.45 M \\
    Trainable Parameters & 0  &  0.80 M & 859.52 M & 55.93 M \\
    Batch Size per GPU   & --  & 12 & 12 & 12\\
    Batch Size    & --  & 192  & 192 & 192 \\
    Iterations    & --  & 250K Steps  & 100K Steps & 250K Steps* \\
    Learning Rate    & --  &  $1\times10^{-4}$ & $1\times10^{-5}$ & $1\times10^{-5}$$\dagger$ \\
    Guidance Scale    & 6  & 6  & 6 & 6 \\
\bottomrule
\end{tabular}

\label{tab:hyperparameter}
\end{table*}

\subsection{Training Details}

We leverage a progressive taxonomy-based approach, where the model learns hierarchical information from higher taxonomic levels, such as \code{Family} and \code{Order}, before specializing in fine-grained traits at the \code{Species} level. This progressive strategy ensures that shared characteristics or traits at higher levels are effectively utilized, providing a strong foundation for generating accurate and semantically aligned species-specific images. This approach significantly enhances the model's ability to capture fine-grained details without the need for full model fine-tuning.

To efficiently train the model, we employ LoRA, which introduces low-rank (4 in all experiments) updates targeted specifically at key attention layers. These updates focus on critical components of the cross-attention mechanism, such as the queries, keys, and values. The LoRA configuration balances the rank of the updates and their scaling factor to ensure they effectively influence the model during training. By using LoRA and our hierarchical conditioning, we only train \(55.93\) million parameters --- approximately \(6.5\%\) of the total model size --- compared to the \(859.52\) million parameters updated during full fine-tuning. This efficiency enables us to achieve results comparable to full fine-tuning while significantly reducing computational overhead.

\begin{table*}[th]
\centering
\caption{Results evaluated at the class level for balanced subsets of the dataset. For classes with fewer than 400 species, all available species are included. We generate additional images per species to maintain balance. For classes with more than 400 species, a subset of 400 species was sampled. We report the FID for visual fidelity, LPIPS for perceptual similarity, and BioCLIP for semantic alignment. 
}
\begin{tabular}{l|cccccc}
\toprule
{Class name}     & {\# of Generated Images} & \multicolumn{1}{l}{{\# of Species}} & {Images per Species} & {FID}  $\downarrow$ & {LPIPS}  $\downarrow$ & {BioCLIP}  $\uparrow$ \\ 
\midrule
{Actinopteri}    & 6000                  & 400                                          & 15                         & 27.75        & 0.7236         & 13.78            \\ 
{Elasmobranchii} & 6000                  & 400                                          & 15                         & 38.67        & 0.7324         & 12.78            \\ 
{Dipneusti}      & 250                   & 5                                            & 50                         & 159.77       & 0.6910         & 8.87             \\ 
{Myxini}         & 1050                  & 21                                           & 50                         & 119.23       & 0.7733         & 11.48            \\ 
{Cladistii}      & 650                   & 13                                           & 50                         & 97.92        & 0.7226         & 8.34             \\ 
{Holocephali}    & 1450                  & 29                                           & 50                         & 109.48       & 0.7364         & 13.10            \\ 
{Petromyzonti}   & 1200                  & 24                                           & 50                         & 93.14        & 0.7954         & 9.91             \\ 
\bottomrule
\end{tabular}

\label{tab:more:result:class:base}
\end{table*}

\section{Additional Results and Analyses}

\subsection{Detailed Results on More Spices}
The results in Table~\ref{tab:more:result:class:base} demonstrate that classes with a higher number of diverse species at the \code{Class} taxonomic level achieve better results. This can be attributed to our hierarchical training approach and the integration of TaxaGuide. By progressively training from higher taxonomy levels, such as \code{Family} and \code{Order}, to lower levels, such as \code{Genus} and \code{Species}, our method effectively transfers shared traits and structural information across related species. This progressive strategy enhances the model's ability to generate detailed and accurate images for classes with richer inter-species diversity.

\mypara{High-Species Diversity Classes.} 
Some classes, such as ``Actinopteri'' and ``Elasmobranchii'', include 400 species, achieved the best FID and BioCLIP scores (\(27.75\) and \(13.78\) for ``Actinopteri''). These results highlight the model's capability to handle diverse species datasets effectively. The relatively low LPIPS values for these classes (\(0.7236\) and \(0.7324\)) indicate high perceptual similarity between generated and real images, suggesting that a large number of species provides sufficient training diversity.

\mypara{Low-Species-Diversity Classes.}
Some classes with fewer species (``Dipneusti'', ``Myxini'', and ``Cladistii'') experienced significantly higher FID values, such as \(159.77\) for ``Dipneusti''. Despite generating 50 images per species to balance these classes, the limited inter-species diversity likely constrains the ability of the model to generalize, resulting in reduced visual fidelity. Similarly, lower BioCLIP scores in these classes (\(8.87\) for \textit{Dipneusti}) indicate weaker semantic alignment, reflecting challenges in capturing fine-grained distinctions when species representation is sparse.

\mypara{Intermediate Classes.} 
Some classes such as ``Holocephali'' and ``Petromyzonti'' with moderate species diversity (29 and 24 species, respectively) showed better FID and BioCLIP scores than ``Dipneusti'' but lagged behind ``Actinopteri'' and ``Elasmobranchii''. The FID scores (\(109.48\) and \(93.14\)) indicate reasonable generation fidelity, while the BioCLIP scores (\(13.10\) and \(9.91\)) suggest effective, though not perfect, semantic alignment. This trend underscores the importance of sufficient inter-species diversity for robust generative performance.

\mypara{Overall Trends.}
The results confirm that \ourmethod performs optimally with a balanced number of species, as seen in the low FID and high BioCLIP scores for \textit{Actinopteri} and \textit{Elasmobranchii}. However, as species diversity decreases, the model struggles to maintain comparable performance, suggesting that enhancing data augmentation or leveraging external sources could further benefit low-diversity classes.

In summary, the analysis highlights the strengths of \ourmethod in generating high-fidelity, semantically aligned images for high-diversity classes and identifies areas for improvement, particularly for classes with limited species representation. Future work could explore advanced balancing techniques or domain adaptation strategies to address these challenges.

\begin{figure}[t]
    \centering
    \includegraphics[width=1\linewidth]{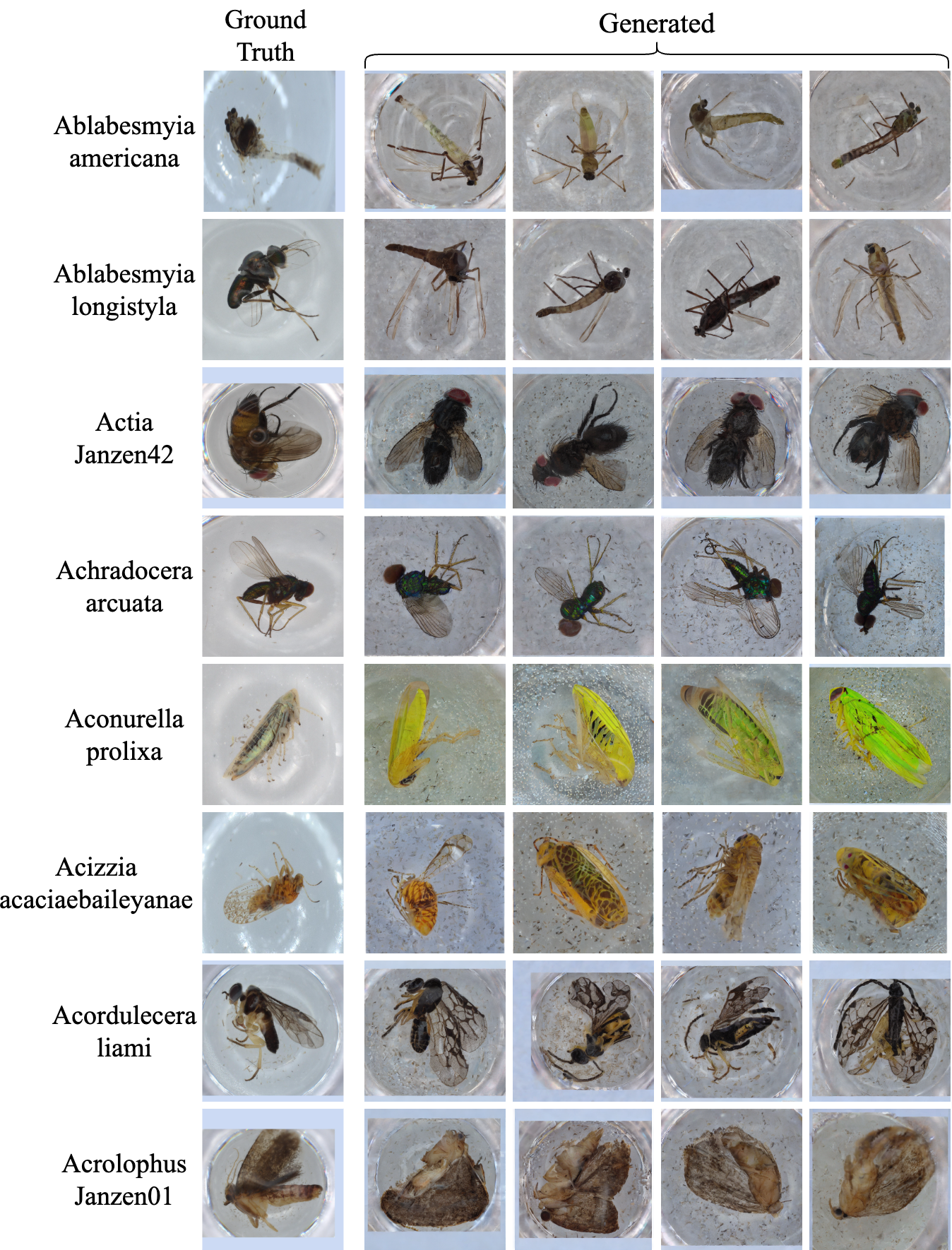}
    \caption{\small \ourmethod results on BIOSCAN-1M dataset~\cite{gharaee2024step}.}
    \vspace{-2mm}
    \label{fig:bio_scan_results}
\end{figure}

\subsection{Results on the BIOSCAN-1M Insect Dataset}

We further validate \ourmethod using the BIOSCAN-1M Insect Dataset \cite{gharaee2024step}, which is a large-scale dataset designed to enable image-based taxonomic classification of insects. The dataset includes high-quality microscope images of insects, each annotated with taxonomic labels ranging from species to higher taxonomic ranks such as genus, family, order, and class. In addition to visual data, the dataset provides associated genetic information, such as DNA barcode sequences and Barcode Index Numbers (BINs), making it a valuable resource for biodiversity assessment and machine learning applications. However, the dataset poses challenges, such as a long-tailed distribution of classes and incomplete taxonomic labeling for many records.

The motivation for using this dataset stems from its potential to validate the capability of our proposed \ourmethod to handle highly diverse and fine-grained taxonomic classes. To ensure consistency with our training pipeline and focus on hierarchical taxonomic generation, we filtered the BIOSCAN-1M dataset to include only records with complete taxonomic information from species to class. After filtering, we retained 84,443 records, ensuring that each sample included annotations for species, genus, family, order, and class levels.

We train \ourmethod on the filtered BIOSCAN-1M dataset for 50 epochs at each taxonomic level, following the same progressive training regime. The progressive training process leverages hierarchical taxonomic relationships, starting from higher levels (e.g., class and order), to capture shared traits before specializing at the species level. This approach allows the model to generate images that are not only visually accurate but also semantically aligned with the hierarchical structure of the taxonomy.

The results, as shown in Figure~\ref{fig:bio_scan_results}, demonstrate the strength of our method in generating high-quality and taxonomically coherent insect images across various levels of the taxonomy. By effectively utilizing hierarchical information and addressing class imbalance through progressive training, \ourmethod excels in fine-grained image generation for a diverse set of insect classes.

\begin{figure}[t]
\centering
\begin{subfigure}{0.48\columnwidth}
  \centering
  \includegraphics[width=1\linewidth]{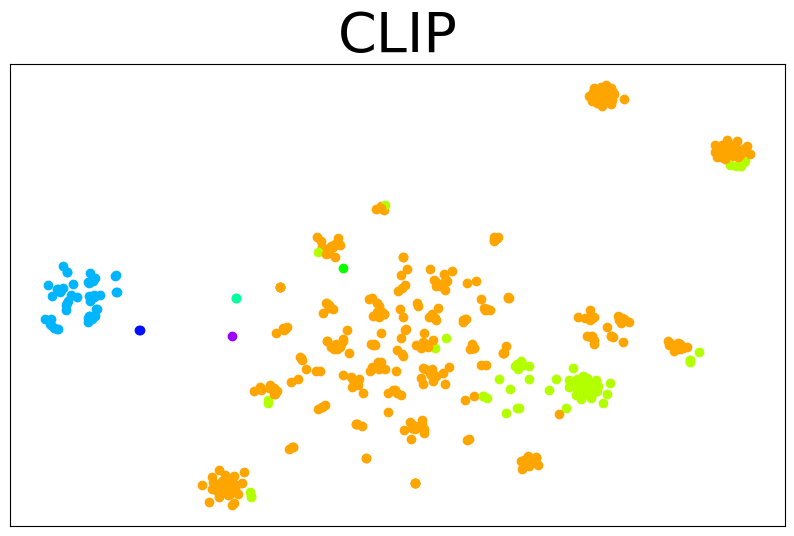}
  \caption{CLIP Embeddings}
\end{subfigure}
\hfill
\begin{subfigure}{0.48\columnwidth}
  \centering
  \includegraphics[width=1\linewidth]{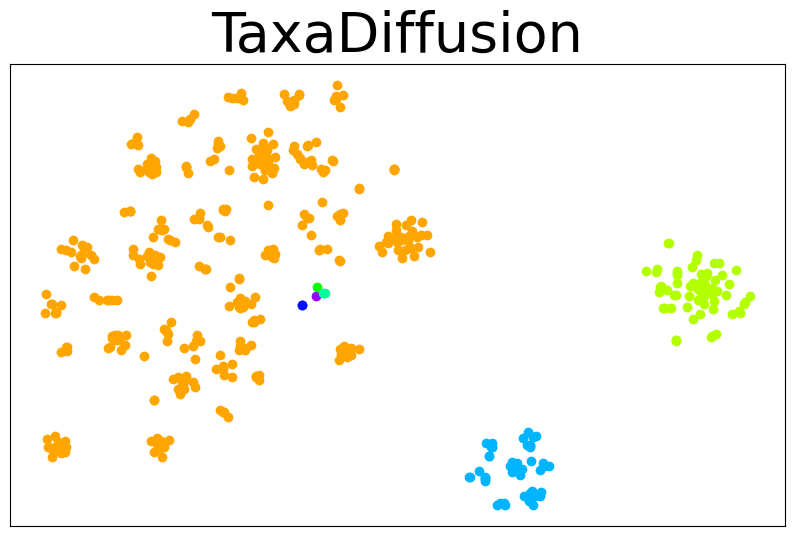}
  \caption{\ourmethod Embeddings}
\end{subfigure}
\vspace{-5pt}
\caption{\textbf{CLIP vs. \ourmethod embeddings.} We show t-SNE visualizations of CLIP and \ourmethod embeddings. Different colors represent different \code{Class} level categories showcasing \ourmethod learns embeddings that are more distinct and form well-separated clusters compared to CLIP embeddings.} 
\label{fig:embeddings}
\vspace{-3mm}
\end{figure}

\begin{figure*}[t]
    \centering
    \includegraphics[width=0.85\linewidth]{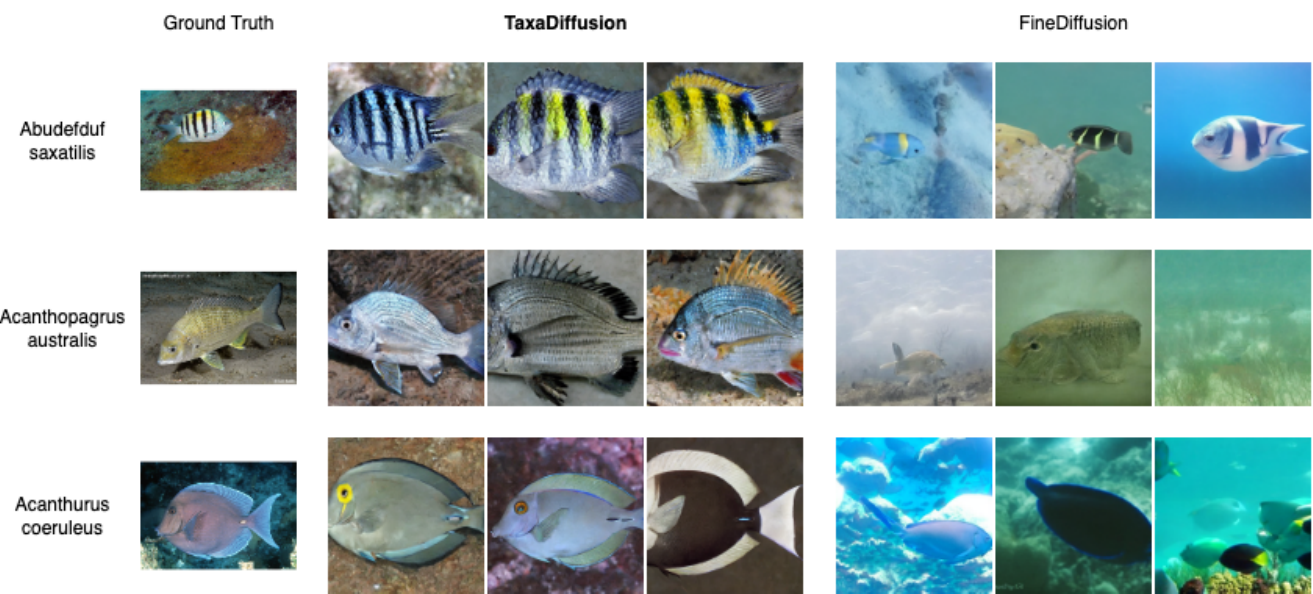}
    \caption{\small \textbf{Comparison with SOTA.} Qualitative comparison between \ourmethod and FineDiffusion.}
    \label{fig:fine_vs_taxa}
\end{figure*}

\subsection{Comparison with State-of-the-Art Methods}

\mypara{Comparison with FineDiffusion} \cite{pan2024finediffusion} Figure \ref{fig:fine_vs_taxa} illustrates the results for a subset of species common to the iNaturalist and FishNet datasets. While FineDiffusion produces similar high-resolution images (512 x 512), the generated images fail to align with ground truth images, lacking species-specific details, whereas, \ourmethod generates images that closely resemble ground truth, effectively capturing fine-grained characteristics unique to each species. This demonstrates the effectiveness of \ourmethod,  achieving superior performance even when compared to FineDiffusion, which utilizes DiT-XL/2 model, a framework shown to outperform U-Net-based architectures \cite{peebles2023scalable}.

\subsection{Embedding Comparison}
We investigate the quality of embeddings of \ourmethod with the CLIP embeddings using t-SNE visualizations. Figure~\ref{fig:embeddings} shows the embeddings at the \code{Family} taxonomic levels for both models with diverse colors representing the unique \code{Class} level. \ourmethod exhibits well-defined clusters, showing a clear separation between different \code{Class} groups, whereas CLIP embeddings used by baselines exhibit overlapping areas, particularly between the orange and green clusters. This hinders its ability to generate images that capture species-specific traits.

\begin{figure*}[t]
    \centering
    \includegraphics[width=0.85\linewidth]{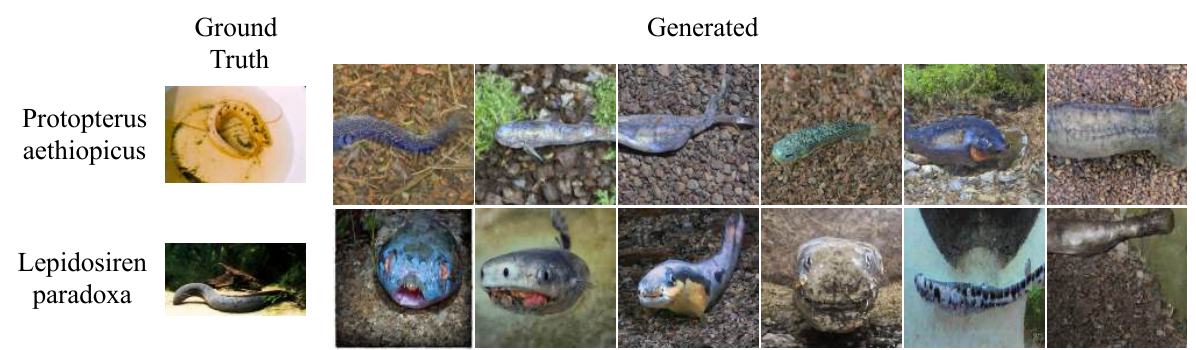}
    \caption{\small \textbf{Failure cases.} Examples of poor generations for the \textit{Dipneusti} class due to insufficient data diversity and representation.}
    \label{fig:supplement_limitation}
\end{figure*}

\subsection{Further Analyses on Guidance Scale}
\label{sup:sec:taxaguide:scale}

\begin{figure*}[ht]
    \centering
    \includegraphics[width=0.95\textwidth]{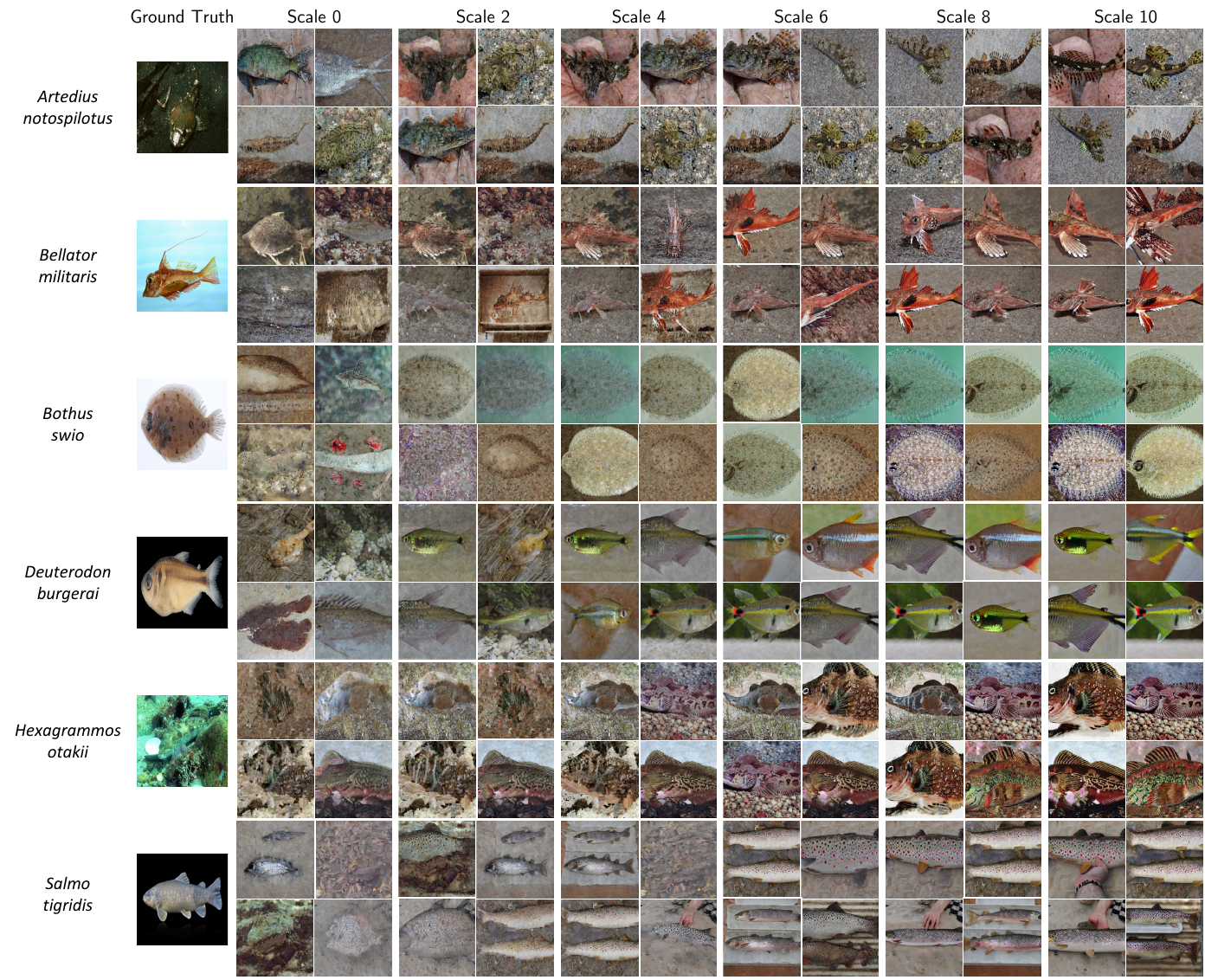}
    \caption{\small \textbf{Scale variations in \ourmethod}. We conduct an analysis of different scale factors used for \ourmethod. Images for each of the six species are generated with 250 inference steps with different \ourmethod scale factors of 0, 2, 4, 6, 8, and 10 to show the balance between the generality and specificity of the generated samples compared to the ground truth training images.}
    \label{fig:guidance_scale}
\end{figure*}

To understand the effect of the guidance scale on image quality and trait specificity, we evaluate our model across a range of guidance scale values: \(0, 2, 4, 6, 8, 10\). The guidance scale directly influences the intensity of the taxonomy-driven conditioning, modulating the balance between the generality and specificity of the generated samples. Figure~\ref{fig:guidance_scale} illustrates representative samples generated using different guidance scales, highlighting the progressive enhancement in fine-grained details and the potential trade-offs at extreme values. 

Our experiments reveal a trend: at lower values (\(0\) and \(2\)), the generated images exhibit limited alignment with the fine-grained taxonomic traits. As the scale increases, the model's focus on species-specific details improves, with noticeable enhancements in morphological accuracy and distinctiveness at a guidance scale of \(6\). Beyond this point, higher values (\(8\) and \(10\)) result in overly restrictive guidance, which limits the diversity of the generated samples and sometimes introduces artifacts due to excessive specificity. Based on these findings, we selected a guidance scale of \(6\) for all subsequent experiments, achieving an optimal balance between fidelity and trait specificity.

\begin{figure}[h]
    \centering
    \includegraphics[width=1\linewidth]{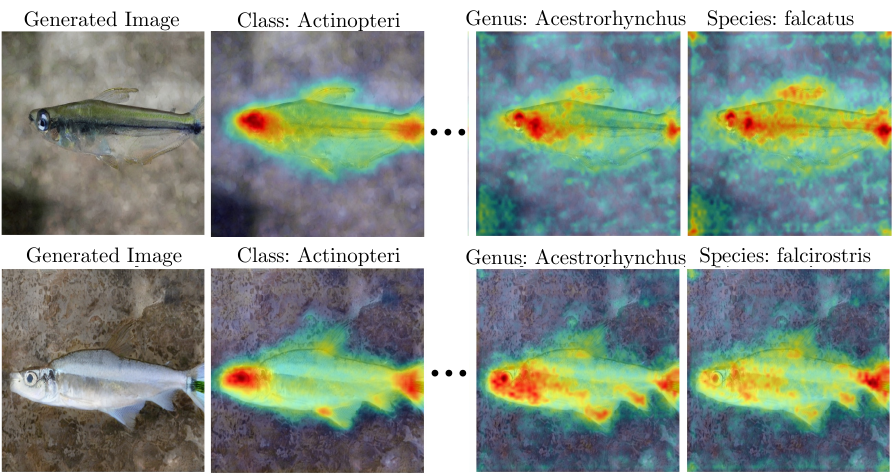}
    \caption{\textbf{Attention maps} of two species \textit{Acestrorhynchus falcatus} and \textit{Acestrorhynchus falcirostris} having the same \code{Genus}}
    \label{fig:attention_maps}
\end{figure}

\subsection{Attention Visualizations}

In Figure \ref{fig:attention_maps}, we show attention maps of each level on two generated fishes \textit{Acestrorhynchus falcatus} and \textit{Acestrorhynchus falcirostris} with the same \code{Genus} but different \code{Species}. Our method captures shared traits at higher levels and distinct features at the \code{Species} level -- the second case focuses more on the tail, less on the head.

\section{Limitations}
While our proposed \ourmethod demonstrates strong generative capabilities across a wide range of taxonomic classes, certain limitations arise in scenarios where the dataset lacks sufficient diversity or representation. One notable case is the class \textit{Dipneusti} within the FishNet dataset. This class contains only 5 species, with a total of 19 samples, which poses a significant challenge for our progressive diffusion framework.

The progressive training approach relies on transferring shared traits and information from higher taxonomic levels to refine the generative process at the species level. However, the extreme sparsity in the \textit{Dipneusti} class limits this transfer, resulting in suboptimal generations for this group. As shown in Figure~\ref{fig:supplement_limitation}, the generated samples for \textit{Dipneusti} often fail to replicate the fine-grained traits and diversity observed in the ground truth. Instead, the model struggles to generalize effectively due to the insufficient data representation, highlighting the importance of adequate class diversity for hierarchical approaches like ours.

Addressing such limitations would require strategies to augment the training process, such as incorporating additional data, leveraging synthetic data generation (using some methods like DreamBooth \cite{ruiz2023dreambooth} to generate images for each specific species), or employing data augmentation techniques. Additionally, methods that enhance learning under extreme class imbalance or sparsity, such as few-shot learning or domain adaptation, could be explored to mitigate these challenges.

Trait discovery is a challenging problem, particularly when working with datasets like FishNet, which predominantly captures images of fish species in natural environments. Many fish species blend into their surroundings through camouflage or appear in diverse orientations, making it difficult to identify traits, especially for species with limited training images. To address these challenges, we plan to incorporate datasets with plain backgrounds, such as those commonly found in museum collections \cite{IDigBio, inhs}, where controlled settings provide clearer views of species traits.

\section{Ethics and Social Impacts}
We focus on advancing generative models to progressively generate images of animal species, aiming to accelerate scientific discovery and research. Our work does not involve sensitive or human data, and we do not foresee any ethical concerns or negative societal impact associated with \ourmethod or the results obtained.

\end{document}